\pdfoutput=1
\documentclass[11pt]{article}

\usepackage{ACL2023}

\usepackage{times}
\usepackage{latexsym}
\usepackage{enumerate}
\usepackage{booktabs}
\usepackage{numprint}
\usepackage{paralist}

\usepackage[T1]{fontenc}
\usepackage[utf8]{inputenc}

\usepackage{microtype}

\usepackage{inconsolata}

\usepackage{amsmath,amssymb}
\usepackage{tikz}
\usepackage{pgfplots}
\usepackage{pgfmath}
\usetikzlibrary{plotmarks,shapes,pgfplots.dateplot}
\usetikzlibrary{math,calc,trees,positioning,arrows,chains,shapes.geometric,%
    decorations.pathreplacing,decorations.pathmorphing,shapes,%
    matrix,shapes.symbols}
\usetikzlibrary{3d,decorations.text,shapes.arrows,fit,backgrounds,quotes,arrows.meta}
\usepackage{ifthen}

\colorlet{color01}{red!10!white}
\colorlet{color02}{orange!10!white}
\colorlet{color03}{green!10!white}
\colorlet{color04}{blue!10!white}
\colorlet{color05}{magenta!20!white}
\colorlet{color06}{black!15!white}
\colorlet{color07}{yellow!10!white}
\colorlet{color08}{white}
\colorlet{color1}{red!80!black}
\colorlet{color1a}{red!50!white}
\colorlet{color2}{orange!80!black}
\colorlet{color2a}{orange!50!white}
\colorlet{color3}{green!50!black}
\colorlet{color3a}{green!90!black}
\colorlet{color4}{blue!80!black}
\colorlet{color4a}{blue!50!white}
\colorlet{color5}{magenta!80!black}
\colorlet{color5a}{magenta!50!white}
\colorlet{color6}{black!80!black}
\colorlet{color6a}{black!50!white}
\colorlet{color7}{white!30!black}
\colorlet{color7a}{black!50!white}
\colorlet{color8}{yellow}
\colorlet{color8a}{black!50!yellow}

\tikzset{every picture/.style={semithick},every path/.style={thick,rounded corners,->}}
\tikzset{
  token/.style={rectangle,rounded corners,draw=black,thick,inner sep=4pt,outer sep=0,minimum width=2em,fill=color03,text centered, minimum height=1.5em,font=\small\sffamily},
  tokenperm/.style={token,fill=color04},
  tokenres/.style={token,fill=color02},
  numperm/.style={token,fill=color08},
  ne/.style={ draw=none, fill=none, font=\footnotesize\sffamily,  minimum height=0em, text centered},
  ncirc/.style={ circle, draw=black, thin, fill=none, font=\footnotesize\sffamily,  minimum height=2em, inner sep=0, text centered},
  nd/.style={ font=\sffamily,  text centered},
  nodecomp/.style={ rectangle,  rounded corners,  draw=black, thick, text width=2em,  font=\footnotesize\sffamily, minimum height=1.3em,  text centered},
  nodevar/.style={ nodecomp, fill=green!10,
  },
  diablo/.style={ rectangle,  rounded corners,  draw=black, thick, text width=10em,  font=\footnotesize\sffamily, minimum height=1em,  text centered},
  branch/.style ={circle,inner sep=0pt,minimum size=1.5mm,fill=black,draw=black},
  diablo2/.style={ rectangle,  rounded corners,  fill=red!10, draw=black!80,thick, text width=3em,  font=\footnotesize\sffamily, minimum height=2.7em,  text centered},
  diafeature/.style={ rectangle, rounded corners=2pt,  fill=green!10, draw=black!80,thick, text width=1em,  font=\footnotesize\sffamily, minimum height=6em,  text centered},
  diafeatnarr/.style={ rectangle, rounded corners=2pt,  fill=green!10, draw=black!80,thick, text width=.5em,  font=\footnotesize\sffamily, minimum height=6em,  text centered},
  dialoss/.style={ diablo2, fill=green!10,
  },
  rotnode/.style={ anchor=center, rotate=90, font=\footnotesize\sffamily
  },
  diaext/.style={ diablo2,fill=yellow!40, },
  diablo3/.style={rectangle, rounded corners, fill=blue!10, draw=blue!40,thick, text width=3.5em,  font=\footnotesize\sffamily\bfseries, text=blue, minimum height=1.5em, text centered},
  line/.style={draw=red,rounded corners,thick, ->, decoration={markings,mark=at position 1 with {\arrow[scale=4,>=stealth]{>}}},postaction={decorate}},
  element/.style={ tape, top color=white, bottom color=blue!50!black!60!, minimum width=8em, draw=blue!40!black!90, very thick, text width=10em, minimum height=3.5em, text centered, on chain},
  every join/.style={->,rounded corners,thick,shorten >=1pt},  decoration={brace},
  lineblue/.style={    join,line width=.07cm,->,blue!20  }
}

\def\RR{\mathbb{R}}
\def\x{\mathbf{x}}
\def\H{\mathbf{H}}
\def\A{\mathbf{A}}
\def\tx{\tilde{\mathbf{x}}}
\def\y{\mathbf{y}}
\def\l{\mathcal{L}}
\def\btheta{\boldsymbol{\theta}}
\def\bpi{\boldsymbol{\pi}}
\newcommand{\brc}[2]{\left(#1\middle| #2\right)}

\newcommand{\pc}[2]{p\brc{#1}{#2}}
\newcommand{\pctl}[2]{{\tilde p}\brc{#1}{#2}}
\newcommand{\pcth}[2]{p_{\btheta}\brc{#1}{#2}}

\newcommand\EEEE[2]{{\mathbb E}_{#1}\left[ #2 \right]}

\newcommand\gr[2]{\nabla_{#2}#1}
\newcommand\ins[1]{\langle\mathrm{INS}_{#1}\rangle}
\def\msk{\langle\mathsf{msk}\rangle}
\def\pad{\langle\mathsf{pad}\rangle}
\def\bos{\langle\mathrm{s}\rangle}
\def\eos{\langle\mathrm{/s}\rangle}

\def\ss{\ensuremath{\langle\mathsf{s}\rangle}\xspace}
\def\sss{\ensuremath{\langle\mathsf{\backslash s}\rangle}\xspace}
\def\instok{\ensuremath{\langle\mathsf{ins}\rangle}\xspace}
\newcommand\inst[1]{\ensuremath{\langle\mathsf{ins}_{#1}\rangle}\xspace}
\def\padtok{\ensuremath{\langle\mathsf{pad}\rangle}\xspace}
\newcommand{\mask}[1]{\ensuremath{\mathsf{msk}_{#1}}\xspace}

\def\model{\ensuremath{\text{GEC-DePenD}}\xspace}
\def\modelb{\ensuremath{\textbf{GEC-DePenD}}\xspace}
\def\sundae{\ensuremath{\text{SUNDAE}}\xspace}
\def\gector{\ensuremath{\text{GECToR}}\xspace}
\def\gectorxlnet{\ensuremath{\text{GECToR}_{\text{XLNet}}}\xspace}
\def\gectorlarge{\ensuremath{\text{GECToR}_{\text{large}}}\xspace}

\def\fhalf{\ensuremath{\mathrm{F}_{0.5}}\xspace}
\def\fhalfb{\ensuremath{\mathbf{F}_{0.5}}\xspace}

\def\subt{_{\mathrm{II}}}
\def\subtt{_{\mathrm{II},\,\mathrm{III}}}
\def\vant{\ensuremath{\text{Vanilla}\subt}\xspace}
\def\sunt{\ensuremath{\sundae\subt}\xspace}
\def\vantt{\ensuremath{\text{Vanilla}\subtt}\xspace}
\def\suntt{\ensuremath{\sundae\subtt}\xspace}

\newcommand{\todo}[1]{[\textbf{TODO}: #1]}

\usepackage[noend]{algorithm2e}
\SetCommentSty{mycommfont}
\SetKwComment{Comment}{/* }{ */}
\RestyleAlgo{ruled}

\title{GEC-DePenD: Non-Autoregressive Grammatical Error Correction with Decoupled Permutation and Decoding}

\author{
	Konstantin Yakovlev\\
	Huawei Noah's Ark Lab\\
	\texttt{yakovlev.konstantin1} \\ \texttt{@huawei-partners.com}\\\And 
 	Alexander Podolskiy\\
	Huawei Noah's Ark Lab\\
	\texttt{podolskiy.alexander} \\ \texttt{@huawei.com}\\\And 
 	Andrey Bout\\
	Huawei Noah's Ark Lab\\
	\texttt{bout.andrey} \\ \texttt{@huawei.com}
	\\\AND
	Sergey Nikolenko\\
	AI Center, NUST MISiS, Moscow, Russia\\
        PDMI RAS, St. Petersburg, Russia\\
	\texttt{sergey@logic.pdmi.ras.ru}
	\\\And
	Irina Piontkovskaya\\
	Huawei Noah's Ark Lab\\
	\texttt{piontkovskaya.irina@huawei.com}
}

\begin{document}
\maketitle
\begin{abstract}
Grammatical error correction (GEC) is an important NLP task that is currently usually solved with autoregressive sequence-to-sequence models. However, approaches of this class are inherently slow due to one-by-one token generation, so non-autoregressive alternatives are needed. In this work, we propose a novel non-autoregressive approach to GEC that decouples the architecture into a permutation network that outputs a self-attention weight matrix that can be used in beam search to find the best permutation of input tokens (with auxiliary \instok{} tokens) and a decoder network based on a step-unrolled denoising autoencoder that fills in specific tokens. This allows us to find the token permutation after only one forward pass of the permutation network, avoiding autoregressive constructions. We show that the resulting network improves over previously known non-autoregressive methods for GEC and reaches the level of autoregressive methods that do not use language-specific synthetic data generation methods. Our results are supported by a comprehensive experimental validation on the ConLL-2014 and Write\&Improve+LOCNESS datasets and an extensive ablation study that supports our architectural and algorithmic choices.
\end{abstract}

\section{Introduction}\label{sec:intro}

Grammatical error correction (GEC) is an important and obviously practically relevant problem in natural language processing. In recent works, GEC has been usually tackled with machine learning approaches, where it has been formalized either as looking for a sequence of edits or transformation tags~\cite{Omelianchuk2020GECToRG} or, more generally, as a sequence-to-sequence text rewriting problem~\cite{naplava-straka-2019-grammatical,grundkiewicz-etal-2019-neural}, a problem that is a natural fit for encoder-decoder architectures.

Latest encoder-decoder architectures indeed define the state of the art in grammatical error correction~\cite{Rothe2021ASR, Lichtarge2020DataWT}. However, the best current results for GEC are achieved by \emph{autoregressive} methods that need to produce output tokens one by one, which significantly hinders inference time and thus limits their applicability in real world solutions. This motivates the development of \emph{non-autoregressive} models that can achieve results similar to autoregressive ones but with a significantly improved runtime. Previously developed non-autoregressive approaches have relied on language-specific transformation tags~\cite{Omelianchuk2020GECToRG,Tarnavskyi2022EnsemblingAK}. In this work, we develop a novel non-autoregressive and language-agnostic approach, called \modelb (\textbf{GEC} with \textbf{De}coupled \textbf{Pe}rmutation \textbf{\&} \textbf{D}ecoding) that yields excellent performance on the GEC task and has other attractive properties. In particular, it is able to output a ranked list of hypotheses that a potential user can choose from.

\begin{figure}[!t]\centering
\resizebox{\linewidth}{!}{
\begin{tikzpicture}[node distance=.2cm]
\def\xx{1.1}
\def\xxx{-.0}
\def\yy{1.}

\foreach[count=\i] \txt in {\ss,I,be,busy,\sss,\instok} {
    \ifthenelse{\i=3}{
        \node[token,fill=color06] (ti\i) at (\i*\xx, 4*\yy) {\txt};
    }{
        \node[token] (ti\i) at (\i*\xx, 4*\yy) {\txt};
    }
}

\foreach[count=\i] \txt in {0,1,2,3,4,5} {
    \ifthenelse{\i=3}{
        \node[numperm,fill=color06] (ni\i) at (\i*\xx, 3*\yy) {\txt};
    }{
        \node[numperm] (ni\i) at (\i*\xx, 3*\yy) {\txt};
    }
}

\foreach[count=\i] \txt in {0,1,5,3,4}
    \node[numperm] (np\i) at (\i*\xx, 2*\yy) {\txt};

\foreach[count=\i] \txt in {\ss,I,\instok,busy,\sss}
    \node[tokenperm] (tp\i) at (\i*\xx, 1*\yy) {\txt};

\foreach[count=\i] \txt in {\ss,I,\mask{1},\mask{2},\mask{3},busy,\sss}
    \node[tokenperm] (te\i) at (\i*\xx, 0*\yy) {\txt};

\foreach[count=\i] \txt in {\ss,I,am,\padtok,\padtok,busy,\sss}
    \node[tokenres] (tr\i) at (\i*\xx, -1*\yy) {\txt};

\foreach \i in {1,2,3,4,5} {
    \draw (ti\i) -- (ni\i);
    \draw (np\i) -- (tp\i);
}

\draw (ti6) -- (ni6);
\draw (ni1) -- (np1);
\draw (ni2) -- (np2);
\draw (ni4) -- (np4);
\draw (ni5) -- (np5);
\draw (ni6) -- ++(0,-.5*\yy) -| (np3);

\foreach \i in {1,2,3} {
    \draw (tp\i) -- (te\i);
}
\draw (tp3) -- ++(0,-.5*\yy) -| (te4);
\draw (tp3) -- ++(0,-.5*\yy) -| (te5);
\draw (tp4) -- ++(0,-.4*\yy) -| (te6);
\draw (tp5) -| (te7);

\foreach \i in {1,2,3,4,5,6,7} {
    \draw (te\i) -- (tr\i);
}

\draw[dashed,rounded corners=5pt] (-3.25*\xx,1.5*\yy) rectangle ++(10.75*\xx,2*\yy);
\node[ne,font=\large\sffamily] at (-1.25*\xx,2.5*\yy) {Permutation network};

\draw[dashed,rounded corners=5pt] (-3.25*\xx,-1.5*\yy) rectangle ++(10.75*\xx,2*\yy);
\node[ne,font=\large\sffamily] at (-1.25*\xx,-.5*\yy) {Decoder network};

% \node[ne] at (6*\xx, 4*\yy) {(a)};
\end{tikzpicture}}

\caption{\model: idea and example.}\label{fig:depend}
\end{figure}
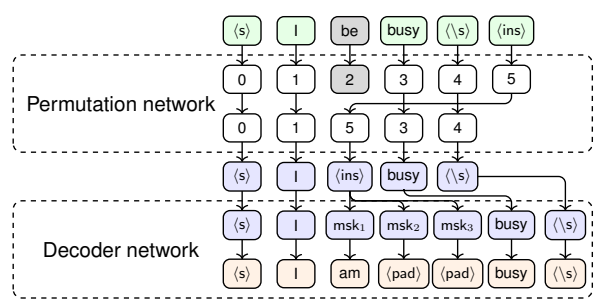

The main idea of \model is to decouple permutation and decoding, with one network producing a permutation of input tokens together with specially added \instok tokens for possible insertions and another network actually infilling \instok tokens. Fig.~\ref{fig:depend} illustrates the idea: the source sentence ``\emph{I be busy}'' is encoded as ``\ss \emph{I be busy} \sss \instok'', the permutation network obtains ``\ss \emph{I} \instok \emph{busy} \sss'', and then the decoder network converts ``\ss \emph{I} \mask{1} \mask{2} \mask{3} \emph{busy} \sss'' into ``\ss \emph{I} \emph{am} \padtok \padtok \emph{busy} \sss'' and outputs ``\emph{I am busy}'' as the corrected sentence. In a single run, the permutation network produces a self-attention matrix for subsequent beam search~\cite{Mallinson2020FELIXFT}, while in the decoder network we use the step-unrolled denoising autoencoder (\sundae) proposed by~\citet{Savinov2022StepunrolledDA}. We also adapt and evaluate several additional techniques including a three-stage training schedule, length normalization, and inference tweaks that improve the final performance.

Thus, our main contributions can be summarized as follows:
\begin{inparaenum}[(i)]
    \item we propose, to the best of our knowledge, the first open-vocabulary iterative non-autoregressive GEC model
    \footnote{
    % We plan to release the source code of our models upon acceptance.
    We release our code at 
    \url{https://github.com/Gibson210/GEC-DePenD}
    }
    based on decoupling permutation and decoding, including
    \item a novel pointing mechanism that can be implemented by a single permutation network without an additional tagger and
    \item a new algorithm for producing ground truth permutations from source (errorful) and target (corrected) sentences, leading to more adequate dataset construction for the GEC task.
\end{inparaenum}
In experimental evaluation, we show that our model outperforms previously known non-autoregressive approaches (apart from \gector that uses language-specific tagging~\cite{Omelianchuk2020GECToRG}) and operates, with similar implementations for backbone networks, several times faster than either autoregressive approaches or \gector.

The paper is organized as follows. Section~\ref{sec:related} surveys related work on both autoregressive and non-autoregressive approaches to GEC. Section~\ref{sec:methods} introduces our approach, including our idea on decoupling permutation and decoding, \sundae, and new ideas for dataset construction and inference tweaks that make our approach work. Section~\ref{sec:eval} shows the main experimental results, Section~\ref{sec:ablation} presents an extensive ablation study that highlights the contributions of various parts of our approach, Section~\ref{sec:concl} concludes the paper, and Section~\ref{sec:limits} discusses the limitations of our approach.

\section{Related work}\label{sec:related}

\noindent\textbf{Synthetic data for grammatical error correction}.
In this work we concentrate on the model part of a GEC pipeline, but we also have to emphasize the importance of data and training pipelines for GEC. We discuss available datasets in Section~\ref{sec:datasets} but it is important to note the role of synthetic data generation for GEC model training. Synthetic data has been used for GEC for a long time~\cite{foster-andersen-2009-generrate,brockett-etal-2006-correcting}, and recent research shows that it can lead to significant performance gains~\cite{stahlberg-kumar-2021-synthetic,htut-tetreault-2019-unbearable}. Approaches for synthetic data generation include character perturbations, dictionary- or edit-distance based replacements, shuffling word order, rule-based suffix transformations, and more~\cite{grundkiewicz-etal-2019-neural,awasthi-etal-2019-parallel,naplava-straka-2019-grammatical,rothe-etal-2021-simple}. However, the most effective methods are language-dependent and require to construct a dictionary of tags and transformations for every language. In particular, \citet{Omelianchuk2020GECToRG} and~\citet{Tarnavskyi2022EnsemblingAK} employ language-specific schemes while we present a language-agnostic approach.

\noindent\textbf{Non-autoregressive machine translation}.
Autoregressive models can be slow due to sequential generation of output tokens. To alleviate this, \citet{Gu2017NonAutoregressiveNM} proposed non-autoregressive generation for machine translation via generating output tokens in parallel. Since non-autoregressive models are not capable of modeling target side dependencies, several approaches have been proposed to alleviate this issue: knowledge distillation~\cite{Gu2017NonAutoregressiveNM, Lee2018DeterministicNN}, iterative decoding~\cite{Ghazvininejad2019MaskPredictPD, Kasai2020NonautoregressiveMT}, latent variables~\cite{DeltaPost, Ma2019FlowSeqNC}, and iterative methods~\cite{Gu2019LevenshteinT, Kasai2020NonautoregressiveMT, Saharia2020NonAutoregressiveMT}.

\noindent\textbf{Autoregressive grammatical error correction}.
Autoregressive models show outstanding performance in the GEC task~\cite{Rothe2021ASR,Lichtarge2020DataWT}. The generation process can be done either in token space~\cite{Lichtarge2020DataWT} or in the space of edits that need to be applied to the source sequence to get the target~\cite{Stahlberg2020Seq2EditsST,Malmi2019EncodeTR}. Using the edit space is motivated by improving the runtime; another way of increasing inference speed is to use aggressive decoding where tokens are generated in parallel and regenerated when there is a difference between source and target sequences~\cite{Sun2021InstantaneousGE}. Combinations with a non-autoregressive error detection model, where an autoregressive decoder generates tokens to be corrected instead of generating the full output sequence, also can improve the running time~\cite{Chen2020ImprovingTE}.

\noindent\textbf{Non-autoregressive text editing models}.
\citet{Mallinson2020FELIXFT} proposed to split the modeling of the target sequence given the source into two parts: the first non-autoregressive model performs tagging and permutes the tokens, and the second model non-autoregressively performs insertions on $\msk$ token positions. In contrast to our work, insertion position are predicted non-autoregressively, which yields lower quality than our approach.
\citet{Omelianchuk2020GECToRG} and~\citet{Tarnavskyi2022EnsemblingAK} proposed to employ a non-autoregressive tagging model for GEC, predicting the transformation of each token. However, these transformations are language-specific, which limits the approach in multilingual settings; in contrast, our approach is language-agnostic. \citet{Awasthi2019ParallelIE}
suggested to construct a language-specific space of all possible edits and proposed iterative refinement that improves decoding performance. They apply the model to the predicted target sequence several times, but this leads to an additional train-test domain shift since the model receives a partially corrected input. In this work we alleviate this issue by using \sundae and perform iterative refinement only with the decoder rather than the entire model, further improving inference speed.

\noindent\textbf{Iterative decoding}. Several approaches were introduced to better capture target-side dependencies. %~\cite{Ghazvininejad2019MaskPredictPD, Savinov2022StepunrolledDA, Lee2018DeterministicNN}.
\citet{Ghazvininejad2019MaskPredictPD} decompose the decoding iteration into two parts: predicting all tokens and masking less confident predictions. \citet{Lee2018DeterministicNN} predict all tokens simultaneously, while \citet{Savinov2022StepunrolledDA} introduce argmax-unrolled decoding that first updates most confident tokens and then less confident ones from the previous iteration.
% ; this approach achieved the best performance.
% , \citet{Savinov2022StepunrolledDA} achieve the best results.

\section{Methods}\label{sec:methods}

\subsection{Decoupling permutation and decoding}

In \model, we separate changes in word order and choosing the actual tokens to insert. Consider a source sentence $\x = (x^1, \ldots, x^n)$ with fixed first and last tokens: $x^1=\ss$, $x^n=\sss$. We append $s$ special tokens responsible for insertions, $\{\inst{i}\}_{i=1}^s$, getting $\tx$, $|\tx|=n+s$. The task is to get an output sequence which is a permutation of a subset of tokens of $\tilde{\mathbf{x}}$, with $\inst{i}$ tokens occurring in order and separated by at least one token from $\x$.
Let $\bpi = \left(\pi^1, \ldots, \pi^p\right)$ be a sequence of indices defining the permutation, with $\pi^1=1$ and $\pi^p=n$ (it points to $\sss$ and indicates stopping). We decompose the architecture according to
\begin{equation}\label{eq:decompose}
\pcth{\y}{\x} = \sum\nolimits_{\bpi}\pcth{\bpi}{\x}\pcth{\y}{\bpi,\x}
\end{equation}
into a \emph{permutation network} implementing $\pcth{\bpi}{\x}$ and a \emph{decoder network} for $\pcth{\y}{\bpi,\x}$ (see Fig.~\ref{fig:depend} for an example). The permutation and decoder networks have a shared encoder, but we do not perform end-to-end training, so in effect we approximate $\sum_{\bpi}$ with a single $\bpi$ (defined in Section~\ref{sec:dataalg}), similar to~\citet{Mallinson2020FELIXFT}.

\textbf{Permutation}. For the permutation network, from the last hidden state of the encoder we obtain a representation 
% of $\tilde{\mathbf{x}}$ from the encoder last hidden state
$\H \in \RR^{(n+s)\times d}$, where $d$ is the latent dimension. We follow \citet{Mallinson2022EdiT5ST} and feed $\H$ through a linear key layer and a single Transformer query layer, obtaining an attention matrix $\A\in \RR^{(n+s)\times (n+s)}$ by computing pairwise dot products of the rows of key and query matrices.
Then the likelihood of the permutation is decomposed as
\begin{multline}\label{eq:pill}
    \log \pc{\bpi}{\A} = \sum\nolimits_{i=2}^p\log \pc{\pi^i}{\bpi^{1:i-1}, \mathbf{A}} = \\
    = \sum\nolimits_{i=2}^p\mathrm{LogSoftmax}(\mathbf{A}_{\pi^{i-1}} + \mathbf{m}_{\pi^{1:i-1}}),
\end{multline}
% \todo{how does $\log \pc{\pi^i}{\pi^{i-1}, \mathbf{A}}$ actually look like? it should be a simple formula rather than a network, right?}
% where $\log \pc{\pi^i}{\bpi^{1:i-1}, \mathbf{A}} = \mathrm{LogSoftmax}(\mathbf{A}_{\pi^{i-1}} + \mathbf{m}_{\pi^{1:i-1}})$,
% and
where $\mathbf{m}_{\pi^{1:i-1}}$ is a mask vector. We mask attention weights in $\A$ in the row $\pi^{i-1}$ for columns $\pi^1,\ldots,\pi^{i-1}$ and do not allow pointing to $\ins{s}$ before $\ins{s-1}$; masking means setting the corresponding ${m}_i$ to $-\infty$. 
%
%$$
%\A = (Q(\H) K(\H)^T.
%$$
%
% \todo{how does $\log \pc{\pi^i}{\pi^{i-1}, \mathbf{A}}$ actually look like? it should be a simple formula rather than a network, right?}
%
The key observation here is that while formula~\eqref{eq:pill} is an autoregressive decomposition for $\bpi$, we do not use it directly during either training or inference. On inference, we get the permutation $\bpi$ with beam search after one encoder pass that gives the attention matrix $\A$ and thus defines $\log \pc{\bpi}{\x}$ for any $\bpi$. Moreover, beam search outputs a ranked list of permutations that can lead to a set of candidate corrections, a feature useful in real world applications.
% , each of which can then be fleshed out by the decoder network, getting a set of candidate corrections for the user.
 % We approximate $$ in~\eqref{eq:decompose} with the sum over beam search outputs; note that for permutations that cannot yield $\y$ we have $\pc{\y}{\bpi,\x}=0$ so in almost all cases the sum reduces to a single $\bpi$.

\textbf{Decoding}. After obtaining $\bpi$, we apply it to the source sentence, getting a permuted input $\bpi(\tx)$, and then apply the decoder network that is supposed to replace $\inst{i}$ in $\bpi(\tx)$ with actual tokens.

During training, the decoder receives a permutation of the source sentence $\tx$ given by an oracle. Following \citet{Mallinson2020FELIXFT}, we replace each $\inst{i}$ token by three $\msk$ tokens (if the target is shorter than 3 tokens we add $\pad$ tokens), sample tokens at $\msk$ positions, and feed the result to the decoder again to calculate the loss function (see Section~\ref{sec:suda} below).
% second term of~\eqref{eq:grad}.

During inference, the decoder iteratively refines tokens at positions where the input had $\msk$ tokens, without any changes to other tokens or their ordering. 
% Formally, define an updated positions as positions of $\msk$ tokens of decoder input. The mask is defined on the first iteration of decoding and is freezed during next iterations. 
We apply the decoder to the output of the previous iteration 
and replace only tokens at positions that were $\msk$ after the permutation (but could change on previous iterations of the decoder).
% and replace tokens at updated positions. Unlike~\citet{Savinov2022StepunrolledDA}, we begin decoding from a sequence of $\msk$ tokens rather than a sequence of uniformly sampled tokens. 
%
To speed up inference, we do not run the decoder if there are no insertions in the prediction.
 % \todo{I don't understand this; unroll is defined as a completely different thing related to loss function construction, why does it matter during inference?}.

\begin{figure*}[!t]
\resizebox{\linewidth}{!}{
\begin{tikzpicture}[node distance=.2cm]
\def\xx{1.5}
\def\yy{1.}
\def\xshift{1.}

\def\inputtext{{"\ss","it","was","20","years","ago","we","were","friends","since","us","were","10","\sss","\inst{1}","\inst{2}","\inst{3}"}}
\def\outputtext{{"\ss","it","was","20","years","ago","and","\padtok","\padtok","we","had","been","\padtok","friends","since","we","\padtok","\padtok","were","10","\sss"}}

\foreach \i in {0,...,16} {
    \node[token] (ti\i) at (\i*\xx, 4.5*\yy) {\pgfmathparse{\inputtext[\i]}\pgfmathresult};
    \ifthenelse{\i=7 \OR \i=10}{
        \node[numperm,fill=color06] (ni\i) at (\i*\xx, 3.5*\yy) {\i};
    }{
        \node[numperm] (ni\i) at (\i*\xx, 3.5*\yy) {\i};
    };
    \draw (ti\i) -- (ni\i);
}

\def\xx{1.6}

\foreach[count=\ii] \i in {0,1,2,3,4,5,14,6,15,8,9,16,11,12,13} {
    \node[numperm] (np\ii) at (\ii*\xx-\xx+\xshift, 2*\yy) {\i};
    \ifthenelse{\i=14}{\draw (ni\i) -- ++(0,-.45*\yy) -| (np\ii);
    }{\ifthenelse{\i=15}{\draw (ni\i) -- ++(0,-.65*\yy) -| (np\ii);
    }{\ifthenelse{\i=16}{\draw (ni\i) -- ++(0,-.85*\yy) -| (np\ii);
    }{\draw (ni\i) -- (np\ii);}}}
    \node[tokenperm] (tp\ii) at (\ii*\xx-\xx+\xshift, 1*\yy) {\pgfmathparse{\inputtext[\i]}\pgfmathresult};
    \draw (np\ii) -- (tp\ii);
}

\def\xx{1.2}

\foreach[count=\ii] \i in {0,1,2,3,4,5,22,22,22,6,22,22,22,8,9,22,22,22,11,12,13} {
    \ifthenelse{\i=22}{}{
        \node[tokenperm] (te\ii) at (\ii*\xx-\xx, -0.5*\yy) {\pgfmathparse{\inputtext[\i]}\pgfmathresult};
    }
}

\foreach \i in {1,2,3} {
    \node[tokenperm] (te7-\i) at (\i*\xx+5*\xx, -0.5*\yy) {\mask{1\i}};
    \node[tokenperm] (te11-\i) at (\i*\xx+9*\xx, -0.5*\yy) {\mask{2\i}};
    \node[tokenperm] (te16-\i) at (\i*\xx+14*\xx, -0.5*\yy) {\mask{3\i}};
}

\foreach \i in {1,2,3,4,5,6} {
    \draw (tp\i) -- (te\i);
}
\draw (tp8) -- (te10);
\draw (tp10) -- (te14);
\draw (tp11) -- (te15);
\foreach \i in {1,2,3} {
    \draw (tp7) -- (te7-\i);
    \draw (tp9) -- (te11-\i);
    \draw (tp12) -- (te16-\i);
}
\draw (tp13) -- (te19);
\draw (tp14) -- (te20);
\draw (tp15) -- (te21);

\foreach[count=\ii] \i in {0,...,20} {
    \node[tokenres] (tr\ii) at (\ii*\xx-\xx, -1.5*\yy) {\pgfmathparse{\outputtext[\i]}\pgfmathresult};
}

\foreach \i in {1,...,6,10,14,15,19,20,21} {
    \draw (te\i) -- (tr\i);
}
\draw (te7-1) -- (tr7);
\draw (te7-2) -- (tr8);
\draw (te7-3) -- (tr9);
\draw (te11-1) -- (tr11);
\draw (te11-2) -- (tr12);
\draw (te11-3) -- (tr13);
\draw (te16-1) -- (tr16);
\draw (te16-2) -- (tr17);
\draw (te16-3) -- (tr18);
\end{tikzpicture}}
%\vspace{-.5cm}

\caption{A complex example of \model with multiple insertions and deletions: ``\emph{It was 20 years ago we were friends since us were 10}'' becomes ``\emph{It was 20 years ago and we had been friends since we were 10}''.}\label{fig:complexex}
\end{figure*}
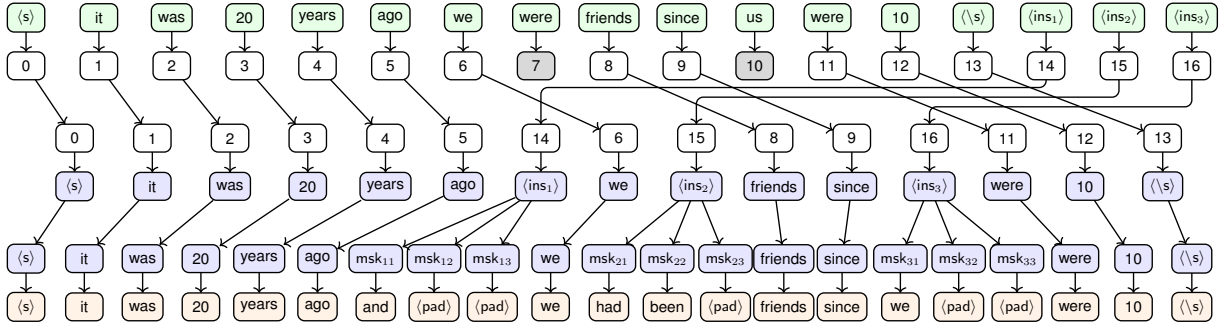

\textbf{Objective}.
% lower bound similarly to 3.1
We minimize the loss function
\begin{equation}\label{eq:loss}
    \mathcal{L}_{\mathrm{total}}(\btheta) = -\lambda_{\mathrm{per}}\log \pcth{\bpi}{\x} - \mathcal{L}_\text{msk}(\btheta),
\end{equation}
where $\mathcal{L}_\text{msk}(\btheta)$ is a lower bound (see Section~\ref{sec:suda}) on the marginal probability of tokens only at $\msk$ positions (the rest are unchanged by the decoder), and $\lambda_{\mathrm{per}}$ is a hyperparameter.
% balancing the permutation and decoding parts of $\mathcal{L}_{\mathrm{total}}$.
%
Fig.~\ref{fig:complexex} shows a complex example of \model operation with multiple insertions.

\subsection{Step-unrolled denoising autoencoder}\label{sec:suda}

For the decoder, we use the step-unrolled denoising autoencoder (\sundae) proposed by~\citet{Savinov2022StepunrolledDA}. Consider a sequence-to-sequence problem with source sequence (sentence) $\x = (x^1, \ldots, x^n)$ and target sequence $\y = (y^1, \ldots, y^m)$. \sundae constructs $T$ intermediate sequences $\y_1,\ldots,\y_T$ with $\y_T=\y$, decomposing

\noindent
\resizebox{\linewidth}{!}{$\pcth{\y_1,\ldots,\y_T}{\x} = \pcth{\y_1}{\x}\prod\limits_{t=2}^T\pcth{\y_t}{\y_{t-1},\x},$}

% Now introduce a sequence of intermediate
% sequences of length $m$: $\{\mathbf{y}_k\}_{k=1}^T, ~\mathbf{y}_T = \mathbf{y}$. Conditional likelihood is written in  the following way:
% \begin{equation}
%     p_{\boldsymbol\theta}(\mathbf{y}_1, \ldots, \mathbf{y}_T | \mathbf{x}) = \prod_{k=2}^T p_{\boldsymbol\theta}(\mathbf{y}_k|\mathbf{y}_{k-1}, \mathbf{x})p_{\boldsymbol\theta}(\mathbf{y}_1|\mathbf{x}),
% \end{equation}
\noindent
where $\btheta$ are model parameters. 
Each term is factorized in a non-autoregressive way, with $y^i_t$ depending only on the previous step $\y_{t-1}$:
\begin{align*}
   \pcth{\y_1}{\x} &= \prod\nolimits_{i=1}^m \pcth{y^i_1}{\x}, \\
    \pcth{\y_t}{\y_{t-1},\x} &= \prod\nolimits_{i=1}^m \pcth{y^i_t}{\y_{t-1},\x},
\end{align*}
so the marginal log-likelihood lower bound is
% takes the following form:
\begin{multline*}
    \log \pcth{\y}{\x} \geq \l(\btheta) = \\ =\EEEE{\y_1,\ldots,\y_{T-1}}{\log\pcth{\y}{\y_{T-1}}}.
\end{multline*}
We follow \citet{Savinov2022StepunrolledDA} and set $T = 2$. The gradient of the lower bound
w.r.t. $\btheta$ is given as
\begin{multline}\label{eq:grad}
\gr{\l(\btheta)}{\btheta} \approx \lambda_0\gr{\log\pcth{\y_1}{\x}}{\btheta}\big|_{\y_1 = \y} + \\
+ (1 - \lambda_0)\EEEE{\y_1}{\gr{\log\pcth{\y}{\y_1,\x}}{\btheta}},
\end{multline}
where $\lambda_0 \in [0, 1]$. \citet{Savinov2022StepunrolledDA} used $\lambda_0 = 0.5$, while we treat $\lambda_0$ as a hyperparameter and optimize it.
This is an approximation since we do not propagate the gradients through sampling $\y_1$. The case $\lambda_0 = 1$ corresponds to $T=1$, i.e. for $\lambda_0 = 1$ target tokens are independent given the source sentence. We call this case \textit{vanilla} below and always perform one decoding step for the vanilla model. If $\lambda_0 \not= 1$, target tokens are dependent given the source; we call this case \sundae.

\subsection{Dataset construction}\label{sec:dataalg}

During training, given source and target sentences $(\x, \y)$, we need to find a permutation $\bpi$ and sequences of tokens that correspond to special $\inst{i}$ tokens. This requires a special algorithm to be applied to available training data; one such algorithm is FELIX proposed by~\citet{Mallinson2020FELIXFT}.

\begin{figure}[!t]\centering\small
\begin{tabular}{c}
\resizebox{.97\linewidth}{!}{
\begin{tikzpicture}[node distance=.2cm]
\def\xx{1.05}
\def\yy{1.}
\def\thinlinewidth{.5pt}
\def\xshift{1.}
\def\inputtext{{"\ss","I","like","films","when","I","was","younger","I","watched","on","TV","\sss"}}
\def\outputtext{{"\ss","I","like","films","I","watched","on","TV","when","I","was","younger","\sss"}}

\foreach \i in {0,1,2} {
    \node[token] (ti\i) at (\i*\xx+3*.3-\i*.3, 4.5*\yy) {\pgfmathparse{\inputtext[\i]}\pgfmathresult};
    \node[numperm] (ni\i) at (\i*\xx+3*.3-\i*.3, 3.5*\yy) {\i};
    \draw[line width=\thinlinewidth] (ti\i) -- (ni\i);
}

\foreach \i in {3,...,12} {
    \node[token] (ti\i) at (\i*\xx, 4.5*\yy) {\pgfmathparse{\inputtext[\i]}\pgfmathresult};
    % \ifthenelse{\i=7 \OR \i=10}{
        % \node[numperm,fill=color06] (ni\i) at (\i*\xx, 3.5*\yy) {\i};
    % }{
    \node[numperm] (ni\i) at (\i*\xx, 3.5*\yy) {\i};
    % };
    \draw[line width=\thinlinewidth] (ti\i) -- (ni\i);
}

\foreach[count=\ii] \i in {0,1,2,3,5,9,10,11,4,8,6,7,12} {
    \ifthenelse{\i=0 \OR \i=1 \OR \i=2}{
        \node[numperm] (np\ii) at (\ii*\xx-\xx+3*.3-\i*.3, 1.75*\yy) {\i};
        \node[tokenperm] (tp\ii) at (\ii*\xx-\xx+3*.3-\i*.3, .75*\yy) {\pgfmathparse{\inputtext[\i]}\pgfmathresult};
    }{
        \node[numperm] (np\ii) at (\ii*\xx-\xx, 1.75*\yy) {\i};
        \node[tokenperm] (tp\ii) at (\ii*\xx-\xx, .75*\yy) {\pgfmathparse{\inputtext[\i]}\pgfmathresult};
    }

    \ifthenelse{\i=9}{\draw[line width=\thinlinewidth] (ni\i) -- ++(0,-.45*\yy) -| (np\ii);
    }{\ifthenelse{\i=10}{\draw[line width=\thinlinewidth] (ni\i) -- ++(0,-.65*\yy) -| (np\ii);
    }{\ifthenelse{\i=11}{\draw[line width=\thinlinewidth] (ni\i) -- ++(0,-.85*\yy) -| (np\ii);
    }{\ifthenelse{\i=5 \OR \i=8}{\draw[line width=1.5pt,color=red!50!black] (ni\i) -- (np\ii);
    }{\draw[line width=\thinlinewidth] (ni\i) -- (np\ii);}}}}
    \draw[line width=\thinlinewidth] (np\ii) -- (tp\ii);
}
\end{tikzpicture}}
\\[6pt] (a) FELIX \\[5pt]
\resizebox{.97\linewidth}{!}{
\begin{tikzpicture}[node distance=.2cm]
\def\xx{1.05}
\def\yy{1.}
\def\thinlinewidth{.5pt}
\def\xshift{1.}
\def\inputtext{{"\ss","I","like","films","when","I","was","younger","I","watched","on","TV","\sss"}}
\def\outputtext{{"\ss","I","like","films","I","watched","on","TV","when","I","was","younger","\sss"}}

\foreach \i in {0,1,2} {
    \node[token] (ti\i) at (\i*\xx+3*.3-\i*.3, 4.5*\yy) {\pgfmathparse{\inputtext[\i]}\pgfmathresult};
    \node[numperm] (ni\i) at (\i*\xx+3*.3-\i*.3, 3.5*\yy) {\i};
    \draw[line width=\thinlinewidth] (ti\i) -- (ni\i);
}

\foreach \i in {3,...,12} {
    \node[token] (ti\i) at (\i*\xx, 4.5*\yy) {\pgfmathparse{\inputtext[\i]}\pgfmathresult};
    % \ifthenelse{\i=7 \OR \i=10}{
        % \node[numperm,fill=color06] (ni\i) at (\i*\xx, 3.5*\yy) {\i};
    % }{
    \node[numperm] (ni\i) at (\i*\xx, 3.5*\yy) {\i};
    % };
    \draw[line width=\thinlinewidth] (ti\i) -- (ni\i);
}

% \def\xx{1.6}
% \foreach[count=\ii] \i in {0,1,2,3,4,9,10,11,12,5,6,7,8,13} {
\foreach[count=\ii] \i in {0,1,2,3,8,9,10,11,4,5,6,7,12} {

    \ifthenelse{\i=0 \OR \i=1 \OR \i=2}{
        \node[numperm] (np\ii) at (\ii*\xx-\xx+3*.3-\i*.3, 1.75*\yy) {\i};
        \node[tokenperm] (tp\ii) at (\ii*\xx-\xx+3*.3-\i*.3, .75*\yy) {\pgfmathparse{\inputtext[\i]}\pgfmathresult};
    }{
        \node[numperm] (np\ii) at (\ii*\xx-\xx, 1.75*\yy) {\i};
        \node[tokenperm] (tp\ii) at (\ii*\xx-\xx, .75*\yy) {\pgfmathparse{\inputtext[\i]}\pgfmathresult};
    }

    \ifthenelse{\i=8}{\draw[line width=1.5pt,color=red!50!black] (ni\i) -- ++(0,-.45*\yy) -| (np\ii);
    }{\ifthenelse{\i=9}{\draw[line width=\thinlinewidth] (ni\i) -- ++(0,-.65*\yy) -| (np\ii);
    }{\ifthenelse{\i=10}{\draw[line width=\thinlinewidth] (ni\i) -- ++(0,-.85*\yy) -| (np\ii);
    }{\ifthenelse{\i=11}{\draw[line width=\thinlinewidth] (ni\i) -- ++(0,-1.05*\yy) -| (np\ii);
    }{\ifthenelse{\i=5}{\draw[line width=1.5pt,color=red!50!black] (ni\i) -- (np\ii);
    }{\draw[line width=\thinlinewidth] (ni\i) -- (np\ii);}}}}}
    \draw[line width=\thinlinewidth] (np\ii) -- (tp\ii);
}

\end{tikzpicture}
}
\\[6pt] (b) Proposed algorithm.
\end{tabular}

\caption{Dataset construction algorithms.} %\vspace{-.5cm}
\label{fig:felix}
\end{figure}
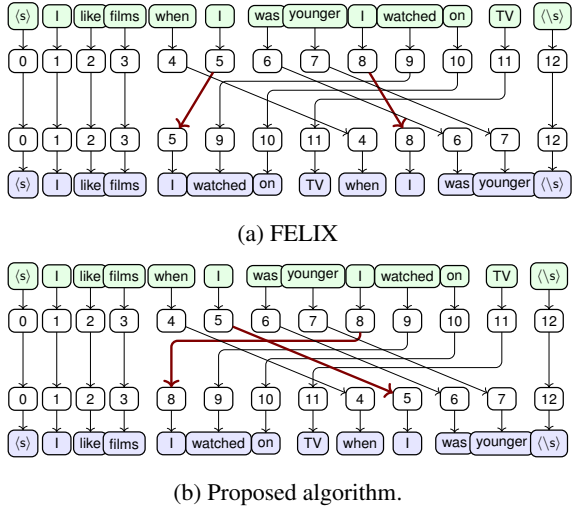

However, we do not use the FELIX dataset construction algorithm
 % by \citet{Mallinson2020FELIXFT, Mallinson2022EdiT5ST} 
because we want to handle cases with repeating tokens 
% when we point to a token corresponding to a different target span 
differently. Fig.~\ref{fig:felix} shows an example: for the input ``\emph{I like films when I was younger I watched on TV}'' the model has to move the clause ``\emph{I watched on TV}'' forward. Both algorithms produce the same tokens but in the permutation, FELIX leaves the ``\emph{I}'' pronouns close to their original locations, breaking the span ``\emph{when I was younger}'', which is undesirable since it makes the permutation network's job harder. 
% Table~\ref{tab:ds_constr}: in this case, instead of inserting token $x_2$, FELIX performs permutation and insertion, while we hypothesize that simplifying the permutation and making decoding a bit harder would lead to better performance since the decoder can be implemented with a more expressive architecture. For instance, in Table~\ref{tab:ds_constr} FELIX unnecessarily breaks a long span that could be left unchanged. \todo{...}

Therefore, we propose a different construction of the permutation $\bpi$ given a source sentence $\mathbf{x}$ and target sentence $\mathbf{y}$. Our algorithm operates as follows:
\begin{enumerate}[(1)]
    \item find all matching spans for the source and target sequences; we iterate over target spans from longer to shorter, and if the current span occurs in the source we remove it from both source and target; at the end of this step, we obtain a sequence of pairs of aligned spans;
    \item reorder source spans and insert missing tokens; we do not allow to reorder spans whose ranks in the target sequence differ by $\ge\mathrm{max\_len}=2$ to make the permutations local; we maximize the total length of spans covered under these constraints with dynamic programming.
\end{enumerate}

%-----Algorithm -------
\begin{algorithm}[!t]\small
% % \begin{figure*}
\caption{Dataset construction}\label{alg:ds_constr}
\KwData{$\mathbf{x}, \mathbf{y}, s, \textrm{max\_len}$}
\KwResult{$\bpi$, dec\_input, dec\_output}
\Comment{List of triples (start\_src, start\_tgt, length)}
aligns = [\,]\;
msk\_x, msk\_y = $\mathbf{x}$, $\mathbf{y}$\;
 \For{len in $\{|\mathbf{y}|, \ldots, 1\}$}{
    \For{i in $\{0, \ldots, |\mathbf{y}| - \mathrm{len} + 1\}$}{
        start = cont\_len(msk\_y[i : i + len], msk\_x)\;
        \If{start != -1} {
            aligns.append(start, i, len)\;
            \Comment{Hide aligned source tokens}
            msk\_x[start : start + len] = -1\;
            \Comment{Hide aligned target tokens}
            msk\_y[i : i + len] = -2\;
        }
    }
}
\Comment{Find the order of appearance of source spans in the target sequence and their lengths}
aligns = sorted(aligns, key=start\_tgt)\;
src\_ranks = argsort(argsort(aligns, key=start\_src))\;
src\_lens = aligns[:, 2]\;
\Comment{Find with dynamic programming a subsequence of src\_ranks s.t. adjacent ranks differ by $\le\text{max\_len}$ with max total length of selected spans; add spans with $\bos$ and $\eos$ manually if not selected}
ids = get\_subsequence(src\_ranks, src\_lens, max\_len)\;
reduced\_aligns = aligns[ids]\;
\Comment{Construct $\bpi$, decoder input, and decoder output}
$\bpi, \text{dec\_output}, \text{dec\_input} = [\,],\ [\,],\ [\,]$\;
last\_src, last\_tgt = -1,\ -1\;
$k = 1$\;
\For{start\_src, start\_tgt, len \textbf{in} reduced\_aligns}{
    \If{\text{last\_tgt} != -1 \textbf{and} $k \leq s$ \textbf{and}\\ \text{start\_tgt} - \text{last\_tgt} $\geq 2$}{
        $\bpi$.append($|\mathbf{x}|$ + k - 1)\;
        k += 1\;
        ins\_seq = $\mathbf{y}$[last\_tgt + 1 : start\_tgt]\;
        ins\_seq.extend([$\pad, \pad$])\;
        dec\_output.extend(ins\_seq[:3])\;
        dec\_input.extend([$\msk$] * 3)\;
    }
    
    $\bpi$.extend([start\_src, \ldots, start\_src + len - 1])\;
    % TODO: "unmask" source and target from -1 and -2
    dec\_input.extend($\mathbf{x}$[start\_src : start\_src + len])\;
    dec\_output.extend($\mathbf{x}$[start\_src : start\_src + len])\;
    last\_tgt = start\_tgt + len - 1\;
    last\_src = start\_src + len - 1\;
}
\end{algorithm}
% \end{figure*}

% \begin{table}[!t]
%     \centering
%     \setlength{\tabcolsep}{3pt}
%     \begin{tabular}{ll}
%         Source & $\bos|x_1|x_3|x_2|x_4|\eos|\ins{1}$ \\
%         Target & $\bos|x_1|x_2|x_3|x_2|x_4|\eos$ \\
%         $\bpi$ (Edit5, FELIX) & $[0, 1, 3, 2, 6, 4, 5]$ \\
%         $\bpi$ (ours) & $[0, 1, 6, 2, 3, 4, 5]$\\
%         \hline
%         % Source & $\bos|However|for|some|rare|diseases|,|text|\eos|\ins{1}$ \\
%         Source & $\bos|I|remember|some|films|when|I|was|younger|I|watched|on|TV|\eos$ \\
%         % Target & $\bos|However|,|for|some|rare|diseases|,|text|\eos|\ins{1}$ \\
%         Target & $\bos|I|remember|some|films|I|watched|on|TV|when|I|was|younger|\eos$ \\
%         % $\bpi$ (Edit5, FELIX) & $[0, 1, 6, 2, 3, 4, 5, 9, 7, 8]$ \\
%         $\bpi$ (FELIX) & $[0, 1, 2, 3, 5, 9, 10, 11, 4, 8, 6, 7, 12]$ \\
%         % $\bpi$ (ours) & $[0, 1, 9, 2, 3, 4, 5, 6, 7, 8]$
%         $\bpi$ (ours) & $[0, 1, 2, 3, 8, 9, 10, 11, 4, 5, 6, 12]$
%     \end{tabular}
%     \caption{Dataset construction comparison. \todo{so what's the difference? looks perfectly equivalent to me}}
%     \label{tab:ds_constr}
% \end{table}

Algorithm~\ref{alg:ds_constr} shows this idea in full formal detail; in the example shown on Fig.~\ref{fig:felix}, it keeps both ``\emph{I}''s with their clauses.
 % Table~\ref{tab:ds_constr} shows that Algorithm~\ref{alg:ds_constr} indeed gives us the desired permutation in this example, and we present an experimental comparison of the FELIX algorithm proposed by \citet{Mallinson2020FELIXFT} with Algorithm~\ref{alg:ds_constr} in Section~\ref{sec:dataabl}.

\subsection{Beam search modifications}\label{sec:inf_tweaks}

To further improve the permutation network, we use two important tricks (see also Section~\ref{sec:ablation}). First, we use \emph{length normalization}, i.e., we divide each candidate score by its length in beam search~\cite{Bahdanau2014NeuralMT,Yang2018BreakingTB}.

Second, we use \emph{inference tweaks} to improve the \fhalf score by rebalancing precision and recall, increasing the former and decreasing the latter~\cite{Omelianchuk2020GECToRG,Tarnavskyi2022EnsemblingAK}. The idea is to make a correction only if we are confident enough. We adopt this idea to beam search decoding in the permutation network. We prioritize the position nearest to the last pointed position on the right. Formally, given a distribution $\pc{\pi^i}{\bpi^{1:i-1}, \mathbf{A}}$, we introduce a \textit{confidence bias} parameter $c \in [0, 1]$ and rescore the distribution as
\begin{multline*}
    \pctl{\pi^i}{\bpi^{1:i-1}, \mathbf{A}} =
    (1-c)\pc{\pi^i}{\bpi^{1:i-1}, \mathbf{A}} + \\
    + c\cdot\mathrm{one\_hot}(\mathrm{right(\bpi^{1:i-1})}),
\end{multline*}
where $\mathrm{right}(\bpi^{1:i-1})$ is the smallest $j\in [\pi^{i-1}+1, n+2]$ such that $j \not\in \bpi^{1:i-1}$.
% $$
%     \mathrm{right}(\bpi^{1:i-1}) =
%     \arg\min_{j \in [\pi^{i-1}+1, |\mathbf{x}| + 1], j \not\in \bpi^{1:i-1}}j.
% $$

% \subsection{Labeling}
% Given a pair of a source and target sentence $(\mathbf{x}, \mathbf{y})$. The task is to find permutation $\boldsymbol\pi$ and sequences of tokens that correspond to special $\mathrm{<INS>}$ tokens. 
% There exists an algorithm proposed in [FELIX, EdiT5] which solves that problem. \todo{???}

% \subsection{Architecture}
% First, write down conditional likelihood:
% \begin{equation}
%     p_{\boldsymbol\theta}(\mathbf{y}|\mathbf{x}) = p_{\boldsymbol\theta}(\boldsymbol\pi|\mathbf{x})\cdot p_{\boldsymbol\theta}(\mathbf{y}|\boldsymbol\pi, \mathbf{x}).
% \end{equation}
% Here $p_{\boldsymbol\theta}(\boldsymbol\pi|\mathbf{x})$ is obtained from permutation  network built on the encoder outputs. The distribution $p_{\boldsymbol\theta}(\mathbf{y}|\boldsymbol\pi, \mathbf{x})$ is modeled in the decoder following step-unrolled denoising concept. 

% During training the decoder receives a permutation of source sentence given by the oracle. Note that each insert token is replaced by 3 $\mathrm{<MASK>}$ tokens. If the corresponding piece of target text is less than 3, we predict $\mathrm{<PAD>}$ token. This is literally infilling strategy proposed in [FELIX]. Then we sample the tokens at $\mathrm{<MASK>}$ positions and feed to the decoder again to calculate another term of \eqref{sundae:grad}. 

\section{Evaluation}\label{sec:eval}

\subsection{Datasets and training stages}\label{sec:datasets}

Each dataset is a parallel corpus of errorful and error-free sentences. Similar to~\cite{Omelianchuk2020GECToRG,Tarnavskyi2022EnsemblingAK,Katsumata2020StrongerBF}, we train \model in three coarse-to-fine training stages.
Table~\ref{tab:datasets} summarizes dataset statistics and which stages of our pipeline they are used on.
For \emph{Stage I} (pretraining), we use the synthetic PIE dataset constructed by~\citet{Awasthi2019ParallelIE} by injecting synthetic grammatical errors into correct sentences.
For training on \emph{Stage II}, we used several datasets:
\begin{inparaenum}[(i)]
\item First Certificate in English (FCE)~\cite{Yannakoudakis2011AND} that contains \numprint{28350} error-coded sentences from English as a second language exams,
\item National University of Singapore Corpus of Learner English (NUCLE)~\cite{Dahlmeier2013BuildingAL} with over 50K annotated sentences from essays of undergraduate students learning English,
\item Write\&Improve+LOCNESS dataset (W\&I+L, also called BEA-2019 in some literature)~\cite{Bryant2019TheBS} intended to represent a wide variety of English levels and abilities, and
\item cLang8~\cite{Rothe2021ASR}, a distilled version of the Lang8 dataset~\cite{Mizumoto2011MiningRL} cleaned with the gT5 model.
\end{inparaenum}
Finally, we used the W\&I+L dataset again for additional training on \emph{Stage III}.

As evaluation data, we used the CoNLL-2014 test dataset~\cite{Ng2014TheCS} with the $M^2$ scorer \cite{Dahlmeier2012BetterEF} and W\&I+L dev and test sets with the ERRANT scorer~\cite{Bryant2017AutomaticAA}. The W\&I+L dev set was used for validation and ablation study; the two test sets, for evaluation.

\subsection{Baseline methods}\label{sec:baselines}

We consider both autoregressive and non-auto\-reg\-ressive baselines.

\textbf{BART}~\cite{lewis-etal-2020-bart} is an autoregressive sequence-to-sequence model; it takes an errorful sentence as input and produces an error-free sentence token by token with the decoder.
We show the scores reported by~\citet{Katsumata2020StrongerBF} and also reimplement the model with a shallow 2-layer decoder (\emph{BART(12+2)} in Table~\ref{tab:results}) and train it according to the stages shown in Section~\ref{sec:datasets}; note that our reimplementation has improved the results.
% \todo{explain greedy and aggressive decoding \cite{Sun2021InstantaneousGE}}. 
We consider two types of decoding: \emph{greedy} and \emph{aggressive greedy}~\cite{Sun2021InstantaneousGE}. In greedy decoding, we generate the token with highest conditional probability. In aggressive greedy decoding, we generate as many tokens as possible in parallel, then re-decode several tokens after the first difference between source and target sequences, and then switch back to aggressive greedy decoding, repeating the procedure until the $\eos$ token. Aggressive greedy decoding is guaranteed to produce the same output as greedy decoding but can be much faster.
For comparison, we also show the state of the art \textbf{T5-XXL} autoregressive model with 11B parameters based on T5~\cite{raffel2020exploring} and trained on a much larger synthetic dataset.

\textbf{FELIX}~\cite{Mallinson2020FELIXFT} is a non-autoregressive model. It consists of two submodels: the first one predicts the permutation of a subset of source tokens and inserts $\msk$ tokens, and the second model infills $\msk$ tokens conditioned on the outputs of the first model. Both stages are done in a non-autoregressive way. Note that the model does not use any language-specific information.

\textbf{Levenshtein Transformer} (LevT)~\cite{Gu2019LevenshteinT, Chen2020ImprovingTE} is a partially non-autoregressive model that does not use language-specific information. It is based on insertions and deletions and performs multiple refinement steps.

\textbf{GECToR}~\cite{Omelianchuk2020GECToRG, Tarnavskyi2022EnsemblingAK} is a non-autoregressive tagging model that uses language-specific information, predicting a transformation for every token. The model is iteratively applied to the corrected sentence from the previous iteration. We compare \gector based on XLNet (\gectorxlnet) and RoBERTa-large (\gectorlarge) pretrained models.

\textbf{Parallel Iterative Edit} (PIE)~\cite{Awasthi2019ParallelIE} is a non-autoregressive model that uses language-specific information. For each source token it predicts the corresponding edits, applying the model iteratively to get the corrected sentence.

\subsection{Experimental setup}

% In this section we present our evaluation results, starting from all details about the experimental setup. 
As the base model for \model we used BART-large~\cite{lewis-etal-2020-bart} with 12 pretrained encoder layers and 2 decoder layers, initialized randomly. The permutation network uses a single Transformer layer, also randomly initialized; the same encoder and decoder configurations were used for our autoregressive baseline BART(12+2).
%
% To show that \sundae allows to better capture inserted token dependencies, we consider vanilla \model and \model with \sundae.
% (see Section~\ref{sec:suda}).
% The main dataset for training is cLang8 \todo{DATA}, and we evaluate the quality of the resulting models on ConLL-2014 and BEA-dev. 
% We considered two baselines: the first does not use unroll to model $\pcth{\mathbf{y}}{\boldsymbol\pi, \mathbf{x}}$, while the second uses unroll. \todo{details? we have much more than two baselines in the tables}

For training we used AdamW~\cite{Loshchilov2017DecoupledWD} with $\beta_1 = 0.9$, $\beta_2=0.999$, $\epsilon=10^{-8}$, weight decay $0.01$, and no gradient accumulation. For stages I and II we used learning rate $3\cdot 10^{-5}$ and constant learning rate scheduler with 500 steps of linear warmup. For stage III we used learning rate $10^{-5}$ and no warmup. For all stages we used $0.1$ dropout, $\text{max\_len}=2$, $s=8$ for Algorithm~\ref{alg:ds_constr}, $\lambda_\text{per}=5$, confidence bias $c\in[0.1, 0.3]$, $2$-$4$ epochs, max 70 tokens per sentence and 3000 tokens per GPU, training on 4 TESLA T4 GPUs.

\begin{table}[!t]
    \centering\setlength{\tabcolsep}{6pt}\small
    % \resizebox{\linewidth}{!}{
    \begin{tabular}{lrrc}
        \toprule
        \textbf{Dataset} & {\textbf{\#sentences}} & {\textbf{\%errorful}} & \textbf{Stages} \\ \midrule
        PIE  & \numprint{9000000} & 100.0 & I \\
        cLang8  & \numprint{2372119} & 57.7 & II \\
        FCE, train  & \numprint{28350} & 62.5 & II \\
        NUCLE  & \numprint{57151} & 37.4 & II\\
        W\&I+L, train  & \numprint{34308} & 66.3 & II, III \\
        \midrule
        W\&I+L, dev  & \numprint{4384} & 64.3 & Val\\
        CoNLL, test  & \numprint{1312} & 71.9 & Test \\
        W\&I+L, test  & \numprint{4477} & N/A & Test \\\bottomrule
        % PIE \cite{Awasthi2019ParallelIE} & \numprint{9000000} & 100.0 & I \\
        % cLang8 \cite{Rothe2021ASR} & \numprint{2372119} & 57.7 & II \\
        % FCE\_train \cite{Yannakoudakis2011AND} & \numprint{28350} & 62.5 & II \\
        % NUCLE \cite{Dahlmeier2013BuildingAL} & \numprint{57151} & 37.4 & II\\
        % W\&I+LOCNESS\_train \cite{Bryant2019TheBS} & \numprint{34308} & 66.3 & II, III \\
        % \hline
        % W\&I+LOCNESS\_dev \cite{Bryant2019TheBS} & \numprint{4384} & 64.3 & Val\\
        % CoNLL\_test \cite{Ng2014TheCS} & \numprint{1312} & 71.9 & Test \\
        % BEA-2019\_test \cite{Bryant2019TheBS} & \numprint{4477} & N/A & Test \\\hline
    \end{tabular}
    % }
    \caption{Training, validation, and test datasets.}
    \label{tab:datasets}
\end{table}

\begin{table*}[!t]
    \centering\small
    % \resizebox{\linewidth}{!}{
    \begin{tabular}{ll|ccc|ccc}
    \toprule
        & & \multicolumn{3}{c|}{\textbf{ConLL-14 test set}} & \multicolumn{3}{c}{\textbf{W\&I+L test set}} \\
        & & \textbf{Prec} & \textbf{Rec} & \fhalfb & \textbf{Prec} & \textbf{Rec} & \fhalfb \\ \midrule
        \multicolumn{5}{c}{\textbf{Autoregressive}} \\ \midrule
        BART-large & \cite{Katsumata2020StrongerBF} & 69.3 & 45.0 & 62.6 & 68.3 & 57.1 & 65.6 \\ 
        BART(12+2) & Our implementation & 69.2 & 49.8 & 64.2 & 69.6 & 63.5 & 68.3 \\
        T5-XXL, 11B parameters & \cite{Rothe2021ASR} & --- & --- & 68.75 & --- & --- & 75.88\\
        \midrule
        % transformer-big, pretrained on synthetic & 69.4 & 42.5 & 61.5 \\
        \multicolumn{5}{c}{\textbf{Non-autoregressive}} \\ \midrule
        LevT & \cite{Chen2020ImprovingTE} & 53.1 & 23.6 & 42.5 & 45.5 & 37.0 & 43.5 \\
        FELIX & \cite{Mallinson2022EdiT5ST} & --- & --- & --- & --- & --- & 63.5 \\ 
        PIE, BERT-large & \cite{Awasthi2019ParallelIE} & 66.1 & 43.0 & 59.7 & 58.0 & 53.1 & 56.9\\
        \gectorlarge, 1 step & \cite{Tarnavskyi2022EnsemblingAK} & 75.4 & 35.3 & 61.4 & 82.03 & 50.81 & 73.05\\
        \gectorlarge, 3 steps & \cite{Tarnavskyi2022EnsemblingAK} & 76.2 & 37.7 & 63.3 & 80.73 & 53.56 & 73.29\\
        \gectorlarge, 5 steps & \cite{Tarnavskyi2022EnsemblingAK} & 76.1 & 37.6 & 63.2 & 80.73 & 53.63 & 73.32\\
        \gectorxlnet & \cite{Omelianchuk2020GECToRG} & 77.5 & 40.1 & 65.3 & 79.2 & 53.9 & 72.4\\
        % \midrule
        % \multicolumn{5}{c}{\textbf{Ours}} \\ \midrule
        \model, vanilla & Ours & 67.8 & 41.3 & 60.1 & 69.5 & 55.3 & 66.1\\
        \model, \sundae & Ours & 73.2 & 37.8 & 61.6 & 72.9 & 53.2 & 67.9\\ \bottomrule
        % \multicolumn{4}{c}{\textbf{Autoregressive}} \\ \midrule
        % BART-large \cite{Katsumata2020StrongerBF} & 69.3 & 45.0 & 62.6 \\ 
        % BART(12+2), ours & 69.2 & 49.8 & 64.2 \\
        % \midrule
        % % transformer-big, pretrained on synthetic & 69.4 & 42.5 & 61.5 \\
        % \multicolumn{4}{c}{\textbf{Non-autoregressive}} \\ \midrule
        % LevT \cite{Chen2020ImprovingTE} & 53.1 & 23.6 & 42.5\\
        % FELIX \cite{Mallinson2020FELIXFT} & - & - & 58.2\\ 
        % PIE, BERT-large \cite{Awasthi2019ParallelIE} & 66.1 & 43.0 & 59.7\\
        % \gectorlarge, 1 step \cite{Tarnavskyi2022EnsemblingAK} & 75.4 & 35.3 & 61.4\\
        % \gectorlarge, 3 steps \cite{Tarnavskyi2022EnsemblingAK} & 76.2 & 37.7 & 63.3\\
        % \gectorlarge, 5 steps \cite{Tarnavskyi2022EnsemblingAK} & 76.1 & 37.6 & 63.2\\
        % \gectorxlnet \cite{Omelianchuk2020GECToRG} & 77.5 & 40.1 & 65.3\\
        % \midrule
        % \multicolumn{4}{c}{\textbf{Ours}} \\ \midrule
        % \model, vanilla & 67.8 & 41.3 & 60.1 \\
        % \model, \sundae, 2 steps & 73.2 & 37.8 & 61.6 \\ \bottomrule
    \end{tabular}
    % }
    \caption{Experimental results on the ConLL-14 and W\&I+L test sets.}\label{tab:results}
\end{table*}

\subsection{Experimental results}\label{sec:results}

The main results of our comparison are presented in Table~\ref{tab:results}. 
We have evaluated the baselines described in Section~\ref{sec:baselines} and \model in two versions: vanilla and \sundae with 2 decoder steps.
The results show that \model outperforms all existing non-autoregressive baselines except for the language-specific \gector family.
% In addition. from  tables \ref{tab:results} and \ref{tab:speed_comp} we see that we outperform all existing non-autoregressive baselines, except language-specifig GECTOR family, in terms of test quality. At the same time, both proposed models outperform GECTOR family in terms of speed.
% \todo{what do we see?}

% \subsection{Performance comparison}

We have also compared GEC baselines and \model in terms of inference speed on the ConLL-2014 test dataset on a single GPU. All models were implemented with the \emph{Transformers} library~\cite{wolf-etal-2020-transformers}. In addition, we do not clip the source sentence, as was done by \citet{Omelianchuk2020GECToRG}, and process one sentence at a time. We used a single TESLA-T4 GPU.
% \todo{let's add brief descriptions of all these models, we have a lot of space to fill yet; somewhere above, this is also needed for the previous section}
% The following models are considered:
% \begin{enumerate}
%     \item Autoregressive BART(12+2) with greedy decoding (BART(12+2), greedy decoding). We trained the model according to our training stages for fair comparison. TODO: add hyperparameters
%     \item Autoregressive BART(12+2) with aggresive decoding \cite{Sun2021InstantaneousGE} (BART(12+2), aggressive decoding). We used the same model weights as in (1).
%     \item GECTOR \cite{Omelianchuk2020GECToRG} with XLNet backbone and performed 5 decoding steps. Note that the results are slightly better than in the paper because we do not clip source sentence if it is too long (GECTOR-base, XLNet, 5 steps).
%     \item GECTOR \cite{Tarnavskyi2022EnsemblingAK} with RoBERTa-large backbone and performed 1, 3, and 5 decoding steps (GECTOR, RoBERTa-large).
%     \item The proposed model without unroll (BART(12+2) w/o unroll).
%     \item The proposed model with unroll and with 3 decoder steps (BART(12+2) w/ unroll, 3 decoder steps).
% \end{enumerate}
Performance results are summarized in Table~\ref{tab:speed_comp}. As we can see, \model outperforms all baselines in terms of inference speed and sets a new standard for performance, running twice faster than even non-autoregressive \gector models. Note that \model with \sundae both outperforms 1-step \gectorlarge in terms of \fhalf on ConLL-14 (Table~\ref{tab:results}) and operates 1.25x faster (Table~\ref{tab:speed_comp}).
The quality gap between \model and its autoregressive counterpart (BART(12+2), our implementation) is reduced but still remains in Table \ref{tab:results}.

\begin{table}[!t]
    \centering\setlength{\tabcolsep}{4pt}\small
    % \resizebox{\linewidth}{!}{
    \begin{tabular}{lrrl}
        \toprule
        \textbf{Model} & {\footnotesize\textbf{Speedup}} & {\footnotesize\textbf{\#params}}\\ \midrule
        % \multicolumn{4}{c}{\textbf{Autoregressive}} \\ \midrule
        BART(12+2), greedy dec. & 1.0x & 238M  \\
        BART(12+2), aggressive dec. & 3.7x & 238M \\ %\midrule
        % \multicolumn{4}{c}{\textbf{Non-autoregressive}} \\ \midrule
        % GECTOR(2020), XLNet, 1 step   & 4.9x & 120M & 63.1 \\
        % GECTOR(2020), XLNet, 3 steps & 2.8x & 120M & 65.6 \\
        \gectorxlnet, 5 steps & 2.8x & 120M \\
        \gectorlarge, 1 step & 3.8x & 360M \\
        \gectorlarge, 3 steps & 2.4x & 360M \\
        \gectorlarge, 5 steps & 2.4x & 360M \\ %\midrule
        % \multicolumn{4}{c}{\textbf{Ours}} \\ \midrule
        \model, vanilla & \textbf{5.3x} & 253M \\
        \model, \sundae & \textbf{4.7x} & 253M \\ 
        % \model, \sundae & 4.1x & 253M \\ 
        \bottomrule
        % \textbf{Model} & {\footnotesize\textbf{Speedup}} & {\footnotesize\textbf{\#Params}} & $\mathbf{F}_{0.5}$\\ \midrule
        % % \multicolumn{4}{c}{\textbf{Autoregressive}} \\ \midrule
        % BART(12+2), greedy dec. & 1.0x & 238M & 64.2 \\
        % BART(12+2), aggressive dec. & 3.7x & 238M & 64.2\\ %\midrule
        % % \multicolumn{4}{c}{\textbf{Non-autoregressive}} \\ \midrule
        % % GECTOR(2020), XLNet, 1 step   & 4.9x & 120M & 63.1 \\
        % % GECTOR(2020), XLNet, 3 steps & 2.8x & 120M & 65.6 \\
        % \gectorxlnet, 5 steps & 2.8x & 120M & \textbf{65.6} \\
        % \gectorlarge, 1 step & 3.8x & 360M & 61.4 \\
        % \gectorlarge, 3 steps & 2.4x & 360M & 63.3 \\
        % \gectorlarge, 5 steps & 2.4x & 360M & 63.2 \\ %\midrule
        % % \multicolumn{4}{c}{\textbf{Ours}} \\ \midrule
        % \model w/o unroll & \textbf{5.3x} & 253M & 60.1 \\
        % \model w/unroll, 3 steps & 4.1x & 253M & 61.6\\ \bottomrule
    \end{tabular}
    % }
    \caption{Performance comparison, ConLL-2014-test.}\label{tab:speed_comp}
     % we report the number of parameters and model speedup over the autoregressive shallow decoder baseline.}\label{tab:speed_comp}
\end{table}

\begin{figure}
	\centering
	\includegraphics[width=0.5\textwidth]{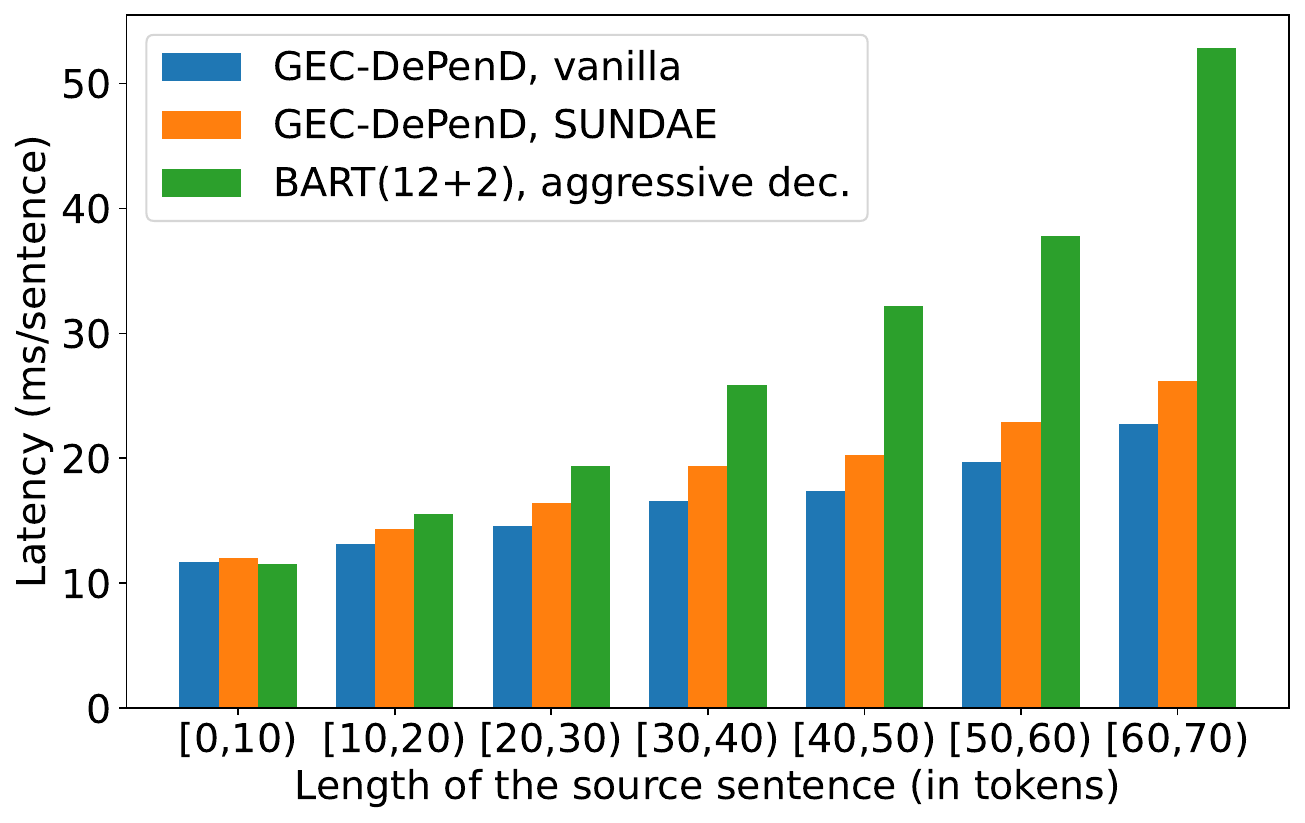}
	\caption{Latency of BART(12+2), aggressive decoding and the proposed family of GEC-DePenD on BEA dev set.}
	\label{fig:latency_bea}
\end{figure}

Figure~\ref{fig:latency_bea} shows a study of the latency with respect to the length of the input sentence in tokens; it shows the results on the BEA-2019 dev set for the proposed GEC-DePenD and autoregressive BART(12+2) with greedy aggressive decoding. We see that the latency of the autoregressive baseline increases faster with increasing input sentence length than for the proposed non-autoregressive models. In addition, the speedup over the autoregressive baseline approaches 2x on sentence lengths from 60 to 70.

\section{Ablation study}\label{sec:ablation}

In this section, we present a detailed ablation study, reporting both ideas that worked (Section~\ref{sec:methods}) and ideas that did not work. Table~\ref{tab:ablation} shows our evaluation on the W\&I+L-dev dataset; below we describe the results of Table~\ref{tab:ablation} from top to bottom. Subscripts (e.g., \vantt) show which training stages were used in the experiment (Section~\ref{sec:datasets}).

% \todo{I wanted to combine the ablation tables but they don't seem to match; e.g., why don't the w/o sinkhorn numbers in Table~\ref{tab:sinkhorn} match any row in Table~\ref{tab:ln}?}

\begin{table}[!t]\centering\small\setlength{\tabcolsep}{4.5pt}
    \begin{tabular}{llll}\toprule
        \textbf{Model} & $\mathbf{Prec}$ & $\mathbf{Rec}$ & \fhalfb \\
        \midrule
        \multicolumn{4}{c}{\textbf{Dataset construction}} \\\midrule
        \vantt + FELIX tagger & 52.5 & 39.5 & 49.3 \\
        \vantt + Algorithm~\ref{alg:ds_constr} & 57.6 & 38.9 & 52.5 \\ \midrule
        \multicolumn{4}{c}{\textbf{Training stages, \sundae and inference tweaks}} \\\midrule
        \vant & 57.9 & 36.5 & 51.8 \\
        \vant + inf. tweaks & 59.3 & 34.6 & 51.9 \\
        \sunt & 56.4 & 39.3 & 51.9 \\
        \sunt + inf. tweaks & 59.9 & 35.0 & 52.4 \\
        \vantt & 54.6 & 42.8 & 51.7 \\
        \vantt + inf. tweaks & 60.6 & 36.5 & 53.5 \\
        \suntt & 54.9 & 43.4 & 52.1 \\
        \suntt + inf. tweaks & 63.5 & 34.3 & 54.3 \\ \midrule
        \multicolumn{4}{c}{\textbf{\sundae hyperparameters selection}} \\\midrule
        1 step, $\lambda_0=0.75$ & 60.8 & 36.5 & 53.6 \\
        1 step, $\lambda_0=0.25$ & 62.9 & 33.9 & 53.7 \\
        1 step, $\lambda_0=0.01$ & 60.8 & 35.8 & 53.4 \\
        2 steps, $\lambda_0=0.75$ & 61.2 & 36.6 & 54.0 \\
        2 steps, $\lambda_0=0.25$ & 63.5 & 34.3 & 54.3 \\
        2 steps, $\lambda_0=0.01$ & 61.6 & 36.4 & 54.1 \\
        3 steps, $\lambda_0 = 0.75$ & 61.3 & 36.7 & 54.0 \\
        3 steps, $\lambda_0 = 0.25$ & 63.5 & 34.3 & 54.3 \\
        3 steps, $\lambda_0 = 0.01$ & 61.7 & 36.4 & 54.1 \\ \midrule
        % \multicolumn{4}{c}{Hypothesis selection} \\\midrule
        \multicolumn{4}{c}{\textbf{Beam search rescoring and sinkhorn}} \\\midrule
        \#1 hypothesis, no length norm & 60.4 & 35.2 & 52.8 \\
        \#2 hypothesis, no length norm & 40.4 & 28.3 & 37.2 \\
        \#3 hypothesis, no length norm & 33.1 & 28.3 & 32.0 \\
        Best of top-3 by GLEU & 71.8 & 45.9 & 64.5 \\ 
        \#1 hypothesis, with length norm & 60.6 & 36.5 & 53.5 \\ 
        % \multicolumn{4}{c}{Rescoring with decoder} \\
        Decoder rescoring, $\lambda_\text{resc} = 0.99$ & 62.3 & 31.8 & 52.3 \\ 
        Decoder rescoring, $\lambda_\text{resc} = 0.999$ & 60.3 & 34.8 & 52.6 \\ 
        Decoder rescoring, $\lambda_\text{resc} = 1$ & 60.4 & 35.2 & 52.8 \\ 
        % \midrule
        % \multicolumn{4}{c}{Sinkhorn layers} \\
        \vantt, 16 sinkhorn layers & 60.6 & 36.7 & 53.6 \\
        \bottomrule
    \end{tabular}
    \caption{Ablation study on W\&I+L-dev.}\label{tab:ablation}
\end{table}

\subsection{Dataset construction}\label{sec:dataabl}

% In this section, we present a detailed ablation study conducted in order to investigate the effects of various design choices in our model. 
First, we show that the proposed dataset construction algorithm (Algorithm~\ref{alg:ds_constr}) indeed yields an increase in performance. We considered the BART-large(12+2) model and performed training without stage I (Section~\ref{sec:datasets}) with FELIX~\cite{Mallinson2020FELIXFT} and Algorithm~\ref{alg:ds_constr}, calibrating the results with inference tweaks. Table~\ref{tab:ablation}
% We report validation \fhalf scores of models calibrated with inference tweaks (see Section~\ref{sec:inf_tweaks}). Table~\ref{tab:ds_ablation} 
shows that the effect from Algorithm~\ref{alg:ds_constr} is positive and significant.

% \begin{table}[!t]
%     \centering
%     \begin{tabular}{cccc}
%     \hline
%         \textbf{Algorithm} & \textbf{Prec} & \textbf{Rec} & $\mathbf{F}_{0.5}$ \\ \hline
%         FELIX & 52.51 & 39.52 & 49.27 \\
%         Proposed & 57.61 & 38.86 & 52.54 \\ \hline
%     \end{tabular}
%     \caption{Effect from the Algorithm~\ref{alg:ds_constr}; the table shows BEA-dev scores.}
%     \label{tab:ds_ablation}
% \end{table}

\subsection{Stage III, \sundae, and inference tweaks}
% In this section we show that incorporating the step-unrolled denoising autoencoder (\sundae) increases the performance.

The next section of Table~\ref{tab:ablation} shows all combinations of two- and three-stage training (Section~\ref{sec:datasets}), vanilla and \sundae model (Section~\ref{sec:suda}), and adding inference tweaks (Section~\ref{sec:inf_tweaks}). We see that each addition---Stage III, \sundae, and inference tweaks---has a positive effect on validation performance in all settings, and the best model, naturally, is \suntt with inference tweaks.
 % that both for two-stage and three-stage training, the model with \sundae shows better validation performance than the vanilla model. In addition to that, for all stages and for all models inference tweaks has positive effect.

% To see the effect of \sundae, we consider the model without the permutation network, where the permutation loss is not optimized during training. This means that instead of predicting $\bpi$ we use an oracle providing the correct permutation $\bpi^*$; this is done to ensure a fair comparison that would not depend on the permutation network's performance. We trained the model on cLang8 and evaluated on ConLL-14. The results are shown in Table~\ref{tab:unroll}; clearly, \sundae increases model performance. Note that the results in Table~\ref{tab:unroll} can also serve as upper bounds on the method's performance (since they are obtained with a perfect permutation network).

\subsection{\sundae hyperparameters}

Next, we show that tuning \sundae hyperparameters, i.e., number of steps and $\lambda_0$ (Section~\ref{sec:suda}), can indeed improve performance; for the final model, we chose $\lambda_0=0.25$ and 2 steps of \sundae.

\subsection{Beam search rescoring and sinkhorn}

We first check how much choosing the right hypothesis from the beam search output will increase the performance. We generate top 3 beam search outputs and use the decoder to fill in $\msk$ tokens. Then we select the hypothesis with the best GLEU score 
\cite{Wu2016GooglesNM} compared to the ground truth, evaluating on W\&I+L-dev. The next section of Table~\ref{tab:ablation} shows that although the results deteriorate significantly from \#1 beam search hypothesis to \#2 and \#3 (suggesting that beam search works as intended), choosing the best out of top three gives a very large increase in the metrics (more than $+0.1$ in terms of the \fhalf measure), so there is a lot of room for improvement in beam search generation. For this improvement, we explored two approaches. First, we tried to rescore hypotheses with decoder scores. Note that the log probability of a hypothesis is the sum of permutation and decoder scores. We introduce $\lambda_\text{resc} \in [0, 1]$ and choose the best hypothesis out of three by the score
$
    % \log \pc{\mathbf{y}}{\mathbf{x}} = 
    \lambda_\text{resc}\log \pc{\bpi}{\mathbf{x}} +
    (1 - \lambda_\text{resc})\log \pc{\mathbf{y}}{\bpi, \mathbf{x}}.
$
We chose the best $\lambda_\text{resc}$ by validation \fhalf but found that while $\lambda_\text{resc}$ does help rebalance precision and recall, the best \fhalf is achieved for $\lambda_\text{resc}^* = 1$, so rescoring with the decoder is not helpful.

The second approach, length normalization (Section~\ref{sec:inf_tweaks}), indeed improved the performance.
% , and we kept it in the final model.

% \subsection{Sinkhorn layers}
Another related idea, the sinkhorn layer, was proposed by \citet{Mena2018LearningLP} as an extension of the Gumbel-Softmax trick and later used for GEC by \citet{Mallinson2022EdiT5ST}. 
% We begin with a brief description of sinkhorn layers. 
For an arbitrary matrix $\mathbf{A}$, a sinkhorn step is defined as follows:
$$\begin{array}{rcl}
    \mathbf{A}' &=& \mathbf{A} - \mathrm{LogSumExp}(\mathbf{A}, \dim=0),\\
    \mathbf{A}^{(1)} &=& \mathbf{A}' - \mathrm{LogSumExp}(\mathbf{A}', \dim=1).
\end{array}$$
$\mathbf{A}^{(1)}$ is the output of the first sinkhorn step, and these steps can be repeated.
% , getting $\mathbf{A}^{(k)}$ after $k$ steps. 
The theoretical motivation here is that when the number of steps $k$ tends to infinity, $\mathrm{exp}(\mathbf{A}^{(k)})$ tends to a doubly stochastic matrix, i.e., after applying $\arg\,\max$ to each row we obtain a valid permutation that does not point to the same token twice; the idea is to make several sinkhorn steps on $\mathbf{A}$ and then optimize the cross-entropy loss as usual.
% then during inference we could simply use $\arg\,\max$ instead of beam search.
% then we still use beam search instead of $\arg\max$ because we need to satisfy the constraints on $\bpi$ (at least that $INS$ is between regular tokens).
% we also used inference tweaks and length normalization.
We have experimented with different variations of sinkhorn layers, but even the best (shown in Table~\ref{tab:ablation}) did not bring any improvements.

\section{Conclusion}\label{sec:concl}

In this work, we have presented \model, a novel method for non-autoregressive grammatical error correction that decouples permutation and decoding steps, adds the step-unrolled denoising autoencoder into the decoder network, changes the dataset construction algorithm to preserve long spans, and uses inference tweaks to improve the results. \model shows the best results among non-autoregressive language-agnostic GEC models and significantly outperforms other models in terms of inference speed. We hope that our approach can become a basis for real life applications of grammatical error correction.
% \section*{Ethics Statement}

\section{Limitations}\label{sec:limits}

The main limitations of our study also provide motivation for future work. 
First, while we have provided an extensive ablation study for \model, there are many more low-level optimizations that can be done to further improve the results. In a real life application, one would be encouraged to investigate these optimizations. Second, obviously, non-autoregressive models, including \model, still lose to state of the art autoregressive models. While the existence of this gap may be inevitable, we believe that it can be significantly reduced in further work.

\section*{Acknowledgements}
We gratefully acknowledge the support of MindSpore, CANN (Compute Architecture for Neural Networks) and Ascend AI Processor used for this research. 

The work of Sergey Nikolenko was prepared in the framework of the strategic project ``Digital Business'' within the Strategic Academic Leadership Program ``Priority 2030'' at NUST MISiS.

% \section*{Acknowledgements}

% \nocite{Homma2020NonAutoregressiveGE, Xu2022FCGECFC}
\bibliography{anthology,custom}

\begin{thebibliography}{44}
\expandafter\ifx\csname natexlab\endcsname\relax\def\natexlab#1{#1}\fi

\bibitem[{Awasthi et~al.(2019{\natexlab{a}})Awasthi, Sarawagi, Goyal, Ghosh,
  and Piratla}]{awasthi-etal-2019-parallel}
Abhijeet Awasthi, Sunita Sarawagi, Rasna Goyal, Sabyasachi Ghosh, and Vihari
  Piratla. 2019{\natexlab{a}}.
\newblock \href {https://doi.org/10.18653/v1/D19-1435} {Parallel iterative edit
  models for local sequence transduction}.
\newblock In \emph{Proceedings of the 2019 Conference on Empirical Methods in
  Natural Language Processing and the 9th International Joint Conference on
  Natural Language Processing (EMNLP-IJCNLP)}, pages 4260--4270, Hong Kong,
  China. Association for Computational Linguistics.

\bibitem[{Awasthi et~al.(2019{\natexlab{b}})Awasthi, Sarawagi, Goyal, Ghosh,
  and Piratla}]{Awasthi2019ParallelIE}
Abhijeet Awasthi, Sunita Sarawagi, Rasna Goyal, Sabyasachi Ghosh, and Vihari
  Piratla. 2019{\natexlab{b}}.
\newblock Parallel iterative edit models for local sequence transduction.
\newblock \emph{ArXiv}, abs/1910.02893.

\bibitem[{Bahdanau et~al.(2014)Bahdanau, Cho, and
  Bengio}]{Bahdanau2014NeuralMT}
Dzmitry Bahdanau, Kyunghyun Cho, and Yoshua Bengio. 2014.
\newblock Neural machine translation by jointly learning to align and
  translate.
\newblock \emph{CoRR}, abs/1409.0473.

\bibitem[{Brockett et~al.(2006)Brockett, Dolan, and
  Gamon}]{brockett-etal-2006-correcting}
Chris Brockett, William~B. Dolan, and Michael Gamon. 2006.
\newblock \href {https://doi.org/10.3115/1220175.1220207} {Correcting {ESL}
  errors using phrasal {SMT} techniques}.
\newblock In \emph{Proceedings of the 21st International Conference on
  Computational Linguistics and 44th Annual Meeting of the Association for
  Computational Linguistics}, pages 249--256, Sydney, Australia. Association
  for Computational Linguistics.

\bibitem[{Bryant et~al.(2019)Bryant, Felice, Andersen, and
  Briscoe}]{Bryant2019TheBS}
Christopher Bryant, Mariano Felice, {\O}istein~E. Andersen, and Ted Briscoe.
  2019.
\newblock The bea-2019 shared task on grammatical error correction.
\newblock In \emph{BEA@ACL}.

\bibitem[{Bryant et~al.(2017)Bryant, Felice, and
  Briscoe}]{Bryant2017AutomaticAA}
Christopher Bryant, Mariano Felice, and Ted Briscoe. 2017.
\newblock Automatic annotation and evaluation of error types for grammatical
  error correction.
\newblock In \emph{Annual Meeting of the Association for Computational
  Linguistics}.

\bibitem[{Chen et~al.(2020)Chen, Ge, Zhang, Wei, and
  Zhou}]{Chen2020ImprovingTE}
Meng~Hui Chen, Tao Ge, Xingxing Zhang, Furu Wei, and M.~Zhou. 2020.
\newblock Improving the efficiency of grammatical error correction with
  erroneous span detection and correction.
\newblock In \emph{Conference on Empirical Methods in Natural Language
  Processing}.

\bibitem[{Dahlmeier and Ng(2012)}]{Dahlmeier2012BetterEF}
Daniel Dahlmeier and Hwee~Tou Ng. 2012.
\newblock Better evaluation for grammatical error correction.
\newblock In \emph{North American Chapter of the Association for Computational
  Linguistics}.

\bibitem[{Dahlmeier et~al.(2013)Dahlmeier, Ng, and
  Wu}]{Dahlmeier2013BuildingAL}
Daniel Dahlmeier, Hwee~Tou Ng, and Siew~Mei Wu. 2013.
\newblock Building a large annotated corpus of learner english: The nus corpus
  of learner english.
\newblock In \emph{BEA@NAACL-HLT}.

\bibitem[{Foster and Andersen(2009)}]{foster-andersen-2009-generrate}
Jennifer Foster and Oistein Andersen. 2009.
\newblock \href {https://aclanthology.org/W09-2112} {{G}en{ERR}ate: Generating
  errors for use in grammatical error detection}.
\newblock In \emph{Proceedings of the Fourth Workshop on Innovative Use of
  {NLP} for Building Educational Applications}, pages 82--90, Boulder,
  Colorado. Association for Computational Linguistics.

\bibitem[{Ghazvininejad et~al.(2019)Ghazvininejad, Levy, Liu, and
  Zettlemoyer}]{Ghazvininejad2019MaskPredictPD}
Marjan Ghazvininejad, Omer Levy, Yinhan Liu, and Luke Zettlemoyer. 2019.
\newblock Mask-predict: Parallel decoding of conditional masked language
  models.
\newblock In \emph{Conference on Empirical Methods in Natural Language
  Processing}.

\bibitem[{Grundkiewicz et~al.(2019)Grundkiewicz, Junczys-Dowmunt, and
  Heafield}]{grundkiewicz-etal-2019-neural}
Roman Grundkiewicz, Marcin Junczys-Dowmunt, and Kenneth Heafield. 2019.
\newblock \href {https://doi.org/10.18653/v1/W19-4427} {Neural grammatical
  error correction systems with unsupervised pre-training on synthetic data}.
\newblock In \emph{Proceedings of the Fourteenth Workshop on Innovative Use of
  NLP for Building Educational Applications}, pages 252--263, Florence, Italy.
  Association for Computational Linguistics.

\bibitem[{Gu et~al.(2017)Gu, Bradbury, Xiong, Li, and
  Socher}]{Gu2017NonAutoregressiveNM}
Jiatao Gu, James Bradbury, Caiming Xiong, Victor O.~K. Li, and Richard Socher.
  2017.
\newblock Non-autoregressive neural machine translation.
\newblock \emph{ArXiv}, abs/1711.02281.

\bibitem[{Gu et~al.(2019)Gu, Wang, and Zhao}]{Gu2019LevenshteinT}
Jiatao Gu, Changhan Wang, and Jake Zhao. 2019.
\newblock Levenshtein transformer.
\newblock In \emph{Neural Information Processing Systems}.

\bibitem[{Htut and Tetreault(2019)}]{htut-tetreault-2019-unbearable}
Phu~Mon Htut and Joel Tetreault. 2019.
\newblock \href {https://doi.org/10.18653/v1/W19-4449} {The unbearable weight
  of generating artificial errors for grammatical error correction}.
\newblock In \emph{Proceedings of the Fourteenth Workshop on Innovative Use of
  NLP for Building Educational Applications}, pages 478--483, Florence, Italy.
  Association for Computational Linguistics.

\bibitem[{Kasai et~al.(2020)Kasai, Cross, Ghazvininejad, and
  Gu}]{Kasai2020NonautoregressiveMT}
Jungo Kasai, James Cross, Marjan Ghazvininejad, and Jiatao Gu. 2020.
\newblock Non-autoregressive machine translation with disentangled context
  transformer.
\newblock In \emph{International Conference on Machine Learning}.

\bibitem[{Katsumata and Komachi(2020)}]{Katsumata2020StrongerBF}
Satoru Katsumata and Mamoru Komachi. 2020.
\newblock Stronger baselines for grammatical error correction using a
  pretrained encoder-decoder model.
\newblock In \emph{AACL}.

\bibitem[{Lee et~al.(2018)Lee, Mansimov, and Cho}]{Lee2018DeterministicNN}
Jason Lee, Elman Mansimov, and Kyunghyun Cho. 2018.
\newblock Deterministic non-autoregressive neural sequence modeling by
  iterative refinement.
\newblock In \emph{Conference on Empirical Methods in Natural Language
  Processing}.

\bibitem[{Lewis et~al.(2020)Lewis, Liu, Goyal, Ghazvininejad, Mohamed, Levy,
  Stoyanov, and Zettlemoyer}]{lewis-etal-2020-bart}
Mike Lewis, Yinhan Liu, Naman Goyal, Marjan Ghazvininejad, Abdelrahman Mohamed,
  Omer Levy, Veselin Stoyanov, and Luke Zettlemoyer. 2020.
\newblock \href {https://doi.org/10.18653/v1/2020.acl-main.703} {{BART}:
  Denoising sequence-to-sequence pre-training for natural language generation,
  translation, and comprehension}.
\newblock In \emph{Proceedings of the 58th Annual Meeting of the Association
  for Computational Linguistics}, pages 7871--7880, Online. Association for
  Computational Linguistics.

\bibitem[{Lichtarge et~al.(2020)Lichtarge, Alberti, and
  Kumar}]{Lichtarge2020DataWT}
Jared Lichtarge, Chris Alberti, and Shankar Kumar. 2020.
\newblock Data weighted training strategies for grammatical error correction.
\newblock \emph{Transactions of the Association for Computational Linguistics},
  8:634--646.

\bibitem[{Loshchilov and Hutter(2017)}]{Loshchilov2017DecoupledWD}
Ilya Loshchilov and Frank Hutter. 2017.
\newblock Decoupled weight decay regularization.
\newblock In \emph{International Conference on Learning Representations}.

\bibitem[{Ma et~al.(2019)Ma, Zhou, Li, Neubig, and Hovy}]{Ma2019FlowSeqNC}
Xuezhe Ma, Chunting Zhou, Xian Li, Graham Neubig, and Eduard~H. Hovy. 2019.
\newblock Flowseq: Non-autoregressive conditional sequence generation with
  generative flow.
\newblock \emph{ArXiv}, abs/1909.02480.

\bibitem[{Mallinson et~al.(2022)Mallinson, Adamek, Malmi, and
  Severyn}]{Mallinson2022EdiT5ST}
Jonathan Mallinson, Jakub Adamek, Eric Malmi, and Aliaksei Severyn. 2022.
\newblock Edit5: Semi-autoregressive text-editing with t5 warm-start.
\newblock \emph{ArXiv}, abs/2205.12209.

\bibitem[{Mallinson et~al.(2020)Mallinson, Severyn, Malmi, and
  Garrido}]{Mallinson2020FELIXFT}
Jonathan Mallinson, Aliaksei Severyn, Eric Malmi, and Guillermo Garrido. 2020.
\newblock Felix: Flexible text editing through tagging and insertion.
\newblock \emph{ArXiv}, abs/2003.10687.

\bibitem[{Malmi et~al.(2019)Malmi, Krause, Rothe, Mirylenka, and
  Severyn}]{Malmi2019EncodeTR}
Eric Malmi, Sebastian Krause, Sascha Rothe, Daniil Mirylenka, and Aliaksei
  Severyn. 2019.
\newblock Encode, tag, realize: High-precision text editing.
\newblock \emph{ArXiv}, abs/1909.01187.

\bibitem[{Mena et~al.(2018)Mena, Belanger, Linderman, and
  Snoek}]{Mena2018LearningLP}
Gonzalo~E. Mena, David Belanger, Scott~W. Linderman, and Jasper Snoek. 2018.
\newblock Learning latent permutations with gumbel-sinkhorn networks.
\newblock \emph{ArXiv}, abs/1802.08665.

\bibitem[{Mizumoto et~al.(2011)Mizumoto, Komachi, Nagata, and
  Matsumoto}]{Mizumoto2011MiningRL}
Tomoya Mizumoto, Mamoru Komachi, Masaaki Nagata, and Yuji Matsumoto. 2011.
\newblock Mining revision log of language learning sns for automated japanese
  error correction of second language learners.
\newblock In \emph{International Joint Conference on Natural Language
  Processing}.

\bibitem[{N{\'a}plava and Straka(2019)}]{naplava-straka-2019-grammatical}
Jakub N{\'a}plava and Milan Straka. 2019.
\newblock \href {https://doi.org/10.18653/v1/D19-5545} {Grammatical error
  correction in low-resource scenarios}.
\newblock In \emph{Proceedings of the 5th Workshop on Noisy User-generated Text
  (W-NUT 2019)}, pages 346--356, Hong Kong, China. Association for
  Computational Linguistics.

\bibitem[{Ng et~al.(2014)Ng, Wu, Briscoe, Hadiwinoto, Susanto, and
  Bryant}]{Ng2014TheCS}
Hwee~Tou Ng, Siew~Mei Wu, Ted Briscoe, Christian Hadiwinoto, Raymond~Hendy
  Susanto, and Christopher Bryant. 2014.
\newblock The conll-2014 shared task on grammatical error correction.

\bibitem[{Omelianchuk et~al.(2020)Omelianchuk, Atrasevych, Chernodub, and
  Skurzhanskyi}]{Omelianchuk2020GECToRG}
Kostiantyn Omelianchuk, Vitaliy Atrasevych, Artem~N. Chernodub, and Oleksandr
  Skurzhanskyi. 2020.
\newblock Gector – grammatical error correction: Tag, not rewrite.
\newblock In \emph{Workshop on Innovative Use of NLP for Building Educational
  Applications}.

\bibitem[{Raffel et~al.(2020)Raffel, Shazeer, Roberts, Lee, Narang, Matena,
  Zhou, Li, Liu et~al.}]{raffel2020exploring}
Colin Raffel, Noam Shazeer, Adam Roberts, Katherine Lee, Sharan Narang, Michael
  Matena, Yanqi Zhou, Wei Li, Peter~J Liu, et~al. 2020.
\newblock Exploring the limits of transfer learning with a unified text-to-text
  transformer.
\newblock \emph{J. Mach. Learn. Res.}, 21(140):1--67.

\bibitem[{Rothe et~al.(2021{\natexlab{a}})Rothe, Mallinson, Malmi, Krause, and
  Severyn}]{Rothe2021ASR}
Sascha Rothe, Jonathan Mallinson, Eric Malmi, Sebastian Krause, and Aliaksei
  Severyn. 2021{\natexlab{a}}.
\newblock A simple recipe for multilingual grammatical error correction.
\newblock In \emph{Annual Meeting of the Association for Computational
  Linguistics}.

\bibitem[{Rothe et~al.(2021{\natexlab{b}})Rothe, Mallinson, Malmi, Krause, and
  Severyn}]{rothe-etal-2021-simple}
Sascha Rothe, Jonathan Mallinson, Eric Malmi, Sebastian Krause, and Aliaksei
  Severyn. 2021{\natexlab{b}}.
\newblock \href {https://doi.org/10.18653/v1/2021.acl-short.89} {A simple
  recipe for multilingual grammatical error correction}.
\newblock In \emph{Proceedings of the 59th Annual Meeting of the Association
  for Computational Linguistics and the 11th International Joint Conference on
  Natural Language Processing (Volume 2: Short Papers)}, pages 702--707,
  Online. Association for Computational Linguistics.

\bibitem[{Saharia et~al.(2020)Saharia, Chan, Saxena, and
  Norouzi}]{Saharia2020NonAutoregressiveMT}
Chitwan Saharia, William Chan, Saurabh Saxena, and Mohammad Norouzi. 2020.
\newblock Non-autoregressive machine translation with latent alignments.
\newblock In \emph{Conference on Empirical Methods in Natural Language
  Processing}.

\bibitem[{Savinov et~al.(2022)Savinov, Chung, Binkowski, Elsen, and van~den
  Oord}]{Savinov2022StepunrolledDA}
Nikolay Savinov, Junyoung Chung, Mikolaj Binkowski, Erich Elsen, and A{\"a}ron
  van~den Oord. 2022.
\newblock Step-unrolled denoising autoencoders for text generation.
\newblock \emph{ArXiv}, abs/2112.06749.

\bibitem[{Shu et~al.(2020)Shu, Nakayama, and Cho}]{DeltaPost}
Raphael Shu, Hideki Nakayama, and Kyunghyun Cho. 2020.
\newblock \href {https://doi.org/10.1609/aaai.v34i05.6413} {Latent-variable
  non-autoregressive neural machine translation with deterministic inference
  using a delta posterior}.
\newblock \emph{Proceedings of the AAAI Conference on Artificial Intelligence},
  34:8846--8853.

\bibitem[{Stahlberg and Kumar(2020)}]{Stahlberg2020Seq2EditsST}
Felix Stahlberg and Shankar Kumar. 2020.
\newblock Seq2edits: Sequence transduction using span-level edit operations.
\newblock \emph{ArXiv}, abs/2009.11136.

\bibitem[{Stahlberg and Kumar(2021)}]{stahlberg-kumar-2021-synthetic}
Felix Stahlberg and Shankar Kumar. 2021.
\newblock \href {https://aclanthology.org/2021.bea-1.4} {Synthetic data
  generation for grammatical error correction with tagged corruption models}.
\newblock In \emph{Proceedings of the 16th Workshop on Innovative Use of NLP
  for Building Educational Applications}, pages 37--47, Online. Association for
  Computational Linguistics.

\bibitem[{Sun et~al.(2021)Sun, Ge, Wei, and Wang}]{Sun2021InstantaneousGE}
Xin Sun, Tao Ge, Furu Wei, and Houfeng Wang. 2021.
\newblock Instantaneous grammatical error correction with shallow aggressive
  decoding.
\newblock \emph{ArXiv}, abs/2106.04970.

\bibitem[{Tarnavskyi et~al.(2022)Tarnavskyi, Chernodub, and
  Omelianchuk}]{Tarnavskyi2022EnsemblingAK}
Maksym Tarnavskyi, Artem~N. Chernodub, and Kostiantyn Omelianchuk. 2022.
\newblock Ensembling and knowledge distilling of large sequence taggers for
  grammatical error correction.
\newblock In \emph{Annual Meeting of the Association for Computational
  Linguistics}.

\bibitem[{Wolf et~al.(2020)Wolf, Debut, Sanh, Chaumond, Delangue, Moi, Cistac,
  Rault, Louf, Funtowicz, Davison, Shleifer, von Platen, Ma, Jernite, Plu, Xu,
  Le~Scao, Gugger, Drame, Lhoest, and Rush}]{wolf-etal-2020-transformers}
Thomas Wolf, Lysandre Debut, Victor Sanh, Julien Chaumond, Clement Delangue,
  Anthony Moi, Pierric Cistac, Tim Rault, Remi Louf, Morgan Funtowicz, Joe
  Davison, Sam Shleifer, Patrick von Platen, Clara Ma, Yacine Jernite, Julien
  Plu, Canwen Xu, Teven Le~Scao, Sylvain Gugger, Mariama Drame, Quentin Lhoest,
  and Alexander Rush. 2020.
\newblock \href {https://doi.org/10.18653/v1/2020.emnlp-demos.6} {Transformers:
  State-of-the-art natural language processing}.
\newblock In \emph{Proceedings of the 2020 Conference on Empirical Methods in
  Natural Language Processing: System Demonstrations}, pages 38--45, Online.
  Association for Computational Linguistics.

\bibitem[{Wu et~al.(2016)Wu, Schuster, Chen, Le, Norouzi, Macherey, Krikun,
  Cao, Gao, Macherey, Klingner, Shah, Johnson, Liu, Kaiser, Gouws, Kato, Kudo,
  Kazawa, Stevens, Kurian, Patil, Wang, Young, Smith, Riesa, Rudnick, Vinyals,
  Corrado, Hughes, and Dean}]{Wu2016GooglesNM}
Yonghui Wu, Mike Schuster, Z.~Chen, Quoc~V. Le, Mohammad Norouzi, Wolfgang
  Macherey, Maxim Krikun, Yuan Cao, Qin Gao, Klaus Macherey, Jeff Klingner,
  Apurva Shah, Melvin Johnson, Xiaobing Liu, Lukasz Kaiser, Stephan Gouws,
  Yoshikiyo Kato, Taku Kudo, Hideto Kazawa, Keith Stevens, George Kurian,
  Nishant Patil, Wei Wang, Cliff Young, Jason~R. Smith, Jason Riesa, Alex
  Rudnick, Oriol Vinyals, Gregory~S. Corrado, Macduff Hughes, and Jeffrey Dean.
  2016.
\newblock Google's neural machine translation system: Bridging the gap between
  human and machine translation.
\newblock \emph{ArXiv}, abs/1609.08144.

\bibitem[{Yang et~al.(2018)Yang, Huang, and Ma}]{Yang2018BreakingTB}
Yilin Yang, Liang Huang, and Mingbo Ma. 2018.
\newblock Breaking the beam search curse: A study of (re-)scoring methods and
  stopping criteria for neural machine translation.
\newblock In \emph{EMNLP}.

\bibitem[{Yannakoudakis et~al.(2011)Yannakoudakis, Briscoe, and
  Medlock}]{Yannakoudakis2011AND}
Helen Yannakoudakis, Ted Briscoe, and Ben Medlock. 2011.
\newblock A new dataset and method for automatically grading esol texts.
\newblock In \emph{Annual Meeting of the Association for Computational
  Linguistics}.

\end{thebibliography}
\bibliographystyle{acl_natbib}

\end{document}